\documentclass{article}

\usepackage[main,final]{neurips_2026}

\usepackage[utf8]{inputenc}
\usepackage[T1]{fontenc}
\usepackage[hidelinks]{hyperref}
\usepackage{url}
\usepackage{graphicx}
\usepackage{booktabs}
\usepackage{amsmath}
\usepackage{amssymb}
\usepackage{mathtools}
\usepackage{subcaption}
\usepackage{multirow}
\usepackage[table]{xcolor}
\usepackage{xspace}
\usepackage{soul}
\usepackage{microtype}
\usepackage[export]{adjustbox}

\definecolor{tabred}{rgb}{1, 0.7, 0.7}        %
\definecolor{taborange}{rgb}{1, 0.85, 0.7}    %
\definecolor{tabyellow}{rgb}{1, 1, 0.7}       %

\usepackage[capitalize,noabbrev]{cleveref}

\usepackage[accsupp]{axessibility}

\newcommand{\parsection}[1]{\noindent\textbf{#1} }

\definecolor{cblue}{rgb}{0, 0.44, 0.75}
\definecolor{cred}{rgb}{0.75, 0, 0}

\newcommand{\methodname}{D\'ej\`aView\xspace}

\newcommand{\z}{\mathbf{z}}
\newcommand{\loss}{\mathcal{L}}

\newcommand{\im}{\mathbf{I}}
\newcommand{\depthmap}{\mathbf{D}}
\newcommand{\raymap}{\mathbf{R}}

\newcommand{\real}{\mathbb{R}\xspace}

\usepackage{makecell}

\setcellgapes{2pt}

\usepackage{pifont}
\definecolor{CheckGreen}{rgb}{0, 0.55, 0}
\definecolor{XRed}{RGB}{180,0,0}
\newcommand{\cmark}{\textcolor{CheckGreen}{\ding{51}}}
\newcommand{\xmark}{\textcolor{XRed}{\ding{55}}}

\usepackage{placeins}

\begin{document}

\title{Déjà View: Looping Transformers for Multi-View 3D Reconstruction}

\author{
\textbf{Alessandro Burzio*\textsuperscript{1,2} \quad
Tobias Fischer*\textsuperscript{1,4} \quad
Sven Elflein\textsuperscript{1,3} \quad
Qunjie Zhou\textsuperscript{1}} \\
\textbf{Riccardo de Lutio\textsuperscript{1} \quad
Jiawei Ren\textsuperscript{1} \quad
Jiahui Huang\textsuperscript{1} \quad
Shengyu Huang\textsuperscript{1}} \\
\textbf{Marc Pollefeys\textsuperscript{4} \quad
Laura Leal-Taix\'e\textsuperscript{1} \quad
Zan Gojcic\textsuperscript{$\dagger$1} \quad
Haithem Turki\textsuperscript{$\dagger$1}} \\
\\
\textsuperscript{1}NVIDIA \quad
\textsuperscript{2}University of Modena and Reggio Emilia, AImageLab  \\
\textsuperscript{3}University of Toronto, Vector Institute \quad
\textsuperscript{4}ETH Z\"urich \\
\textbf{*}Equal contribution \quad
\textsuperscript{$\dagger$}Equal supervision
}

\maketitle

\begin{figure}[!ht]
  \centering
  \includegraphics[width=\linewidth]{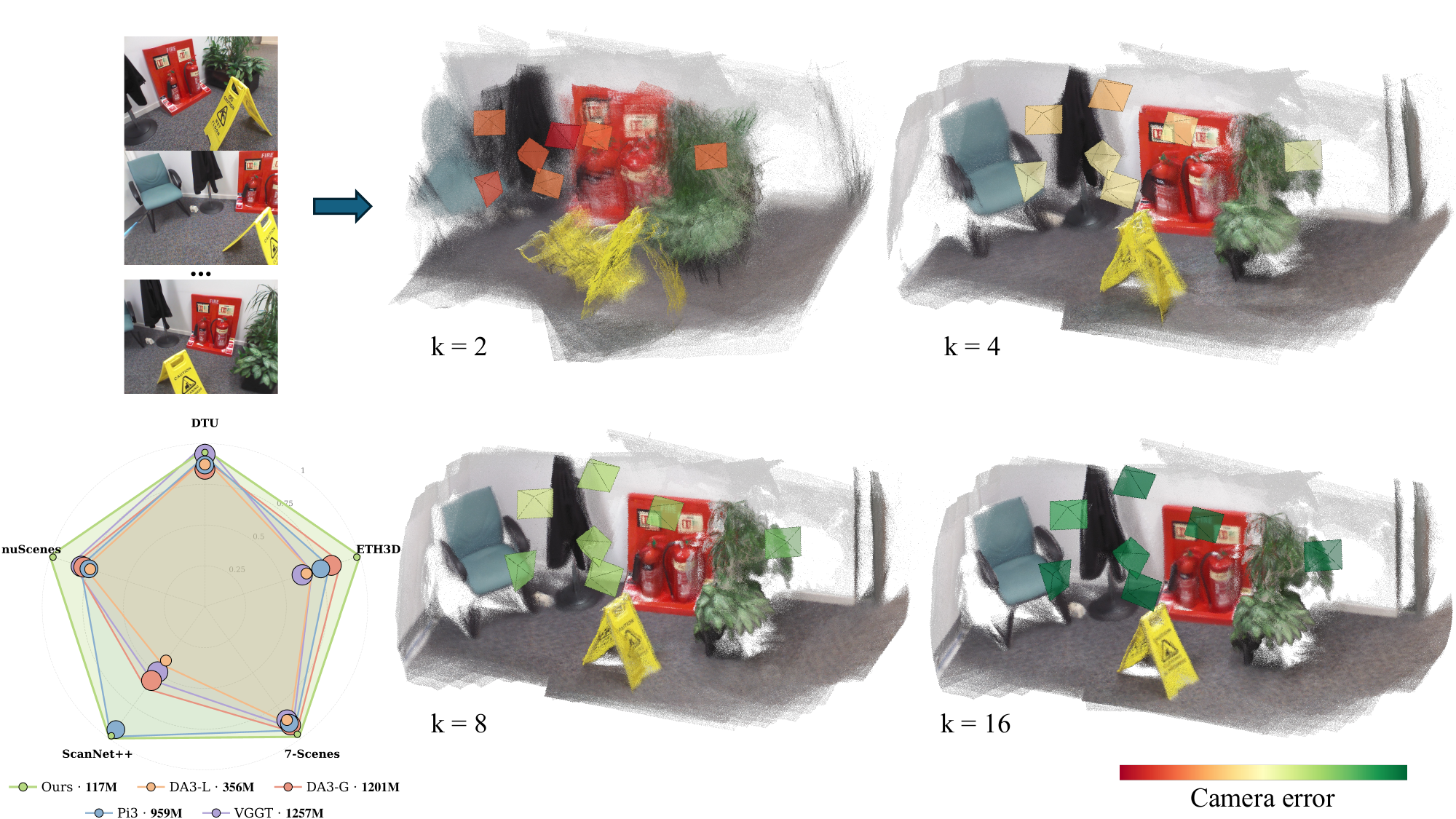}
  \caption{
    \textbf{\methodname.} Given multiple input views \textbf{(top-left)}, \methodname{} reconstructs camera poses and consistent depth by repeatedly applying the \emph{same} transformer block, with the number of refinement steps $K$ exposed as an inference-time compute knob. Decoding the intermediate state of a single $K{=}16$ forward pass at iterations $k \in \{2, 4, 8, 16\}$ shows progressively sharper geometry and more accurate camera poses (\textbf{right}; frustums are colored by per-camera error). Across five benchmarks \textbf{(bottom-left)}, \methodname{} matches or surpasses much larger feed-forward baselines at a small fraction of their parameter count (dot area).
  }
  \label{fig:teaser}
\end{figure}

\begin{abstract}
Recent feed-forward 3D reconstruction transformers have scaled to over a billion parameters, following the broader trend of increasing model capacity in computer vision. Yet emerging evidence suggests that contiguous transformer layers often behave like repeated applications of similar operations~\citep{jacobs2025raptor}, and multi-view reconstruction transformers refine their predictions progressively across decoder depth~\citep{stary2025understanding}. We posit that model depth partially buys iteration, paid for inefficiently in unique parameters, and instead make that iteration explicit in architecture. Our model, \methodname, applies a single looped transformer block recurrently to per-view features for $K$ refinement steps. Trained once, it exposes $K$ as an inference-time compute knob within its trained range, matching or outperforming substantially larger feed-forward baselines across five reconstruction benchmarks spanning indoor, outdoor, object-centric, and driving scenes, while using a fraction of their parameters and comparable or lower compute. Importantly, the same looped block formulation outperforms an otherwise identical variant with independent per-step parameters under matched training data and compute, suggesting that explicit iteration is not merely a compute-efficient substitute for capacity but a stronger inductive bias for multi-view 3D reconstruction. We release code and model at \href{https://research.nvidia.com/labs/dvl/projects/dvlt/}{https://research.nvidia.com/labs/dvl/projects/dvlt/}.
\end{abstract}

\section{Introduction}
\label{sec:intro}

Recovering 3D structure from images has traditionally relied on a Structure-from-Motion (SfM) pipeline~\citep{schoenberger2016sfm,pan2024global}. These systems decompose reconstruction into feature extraction and matching, pose estimation, triangulation, and global bundle adjustment, yielding an inherently iterative process that alternates between registering new views and re-adjusting previous estimates.

More recently, feed-forward methods~\citep{dust3r,leroy2025grounding,vggt,lin2025depth,wang2025pi,keetha2026mapanything,vggt_omega} have replaced this pipeline with an end-to-end transformer that regresses 3D geometry from images in a single forward pass.
Their gains, in line with the broader trend in computer vision~\citep{dosovitskiy2021image,zhai2022scaling,radford2021clip,oquab2023dinov2,kirillov2023sam}, have largely been driven by scaling model capacity through backbone depth and width~\citep{lin2025depth}.

Yet iteration may not have disappeared at all --- it may simply have been absorbed into network depth.
\citet{jacobs2025raptor} show that for some applications the $L$ layers of a Vision Transformer (ViT) can be replaced with $K \ll L$ recurrent applications of a looped block with little loss in accuracy.
\citet{stary2025understanding} further probe the decoder of DUSt3R~\citep{dust3r} layer-by-layer and find that its pointmap predictions are themselves iteratively refined across depth, despite the layers being independently parameterized.

Together, these observations suggest that part of the benefit of depth in modern reconstruction transformers arises from implicit iterative refinement, at the cost of redundant layer-specific parameters.

We therefore make this iterative process explicit in the architecture, rather than relying on model depth to realize it implicitly. Starting from per-view DINOv2~\citep{oquab2023dinov2} features, we apply a single looped transformer block recurrently for $K$ refinement steps. By sampling $K$ from $[K_\text{min}, K_\text{max}]$ during training, a single checkpoint exposes $K$ as an inference-time compute knob without retraining. Analyzing the trained recurrence reveals that it does not converge to a fixed point. Instead, each step progressively aligns the state's direction toward its endpoint, a regime we call \emph{directional refinement}.

The resulting model, \methodname, matches or surpasses much larger feed-forward baselines across five reconstruction benchmarks spanning indoor, outdoor, object-centric, and driving scenes, at a small fraction of their parameters and comparable or lower compute. Importantly, we show that our looping formulation with shared weights significantly outperforms an otherwise identical variant with independent per-step parameters under the same training data and compute budget. We take this as evidence that iterative refinement with shared weights is a viable alternative to parameter scaling for 3D reconstruction.

We summarize our contributions as follows:
\begin{itemize}
    \item \methodname, a looping transformer for multi-view 3D reconstruction that applies a single shared block recurrently to a DINOv2-initialized state, with each step conditioned on a continuous time interval.
    \item Evidence that weight sharing improves reconstruction under matched training data and compute, together with analyses of directional refinement and increasingly accurate correspondence search across iterations.
    \item A single checkpoint that serves multiple inference budgets at near-specialist quality within the trained range, while achieving strong reconstruction results across five benchmarks with substantially fewer parameters than the evaluated baselines.
\end{itemize}

\section{Related Work}
\label{sec:related_work}

Our work draws on several lines of research spanning 3D reconstruction, iterative refinement for geometry estimation, and weight-tied network design.

\parsection{Multi-view 3D reconstruction.}
Classical Structure-from-Motion~\citep{schoenberger2016sfm,pan2024global} recovers geometry through feature matching, pose estimation, and bundle adjustment, but is brittle on in-the-wild scenes with weak texture or dynamic content. Learning-based methods have progressively replaced individual stages of this pipeline, from multi-view stereo~\citep{yao2018mvsnet} to fully end-to-end systems. DUSt3R~\citep{dust3r} and MASt3R~\citep{leroy2025grounding} regress pairwise pointmaps from a CroCo-pretrained~\citep{weinzaepfel2022croco} backbone, while VGGT~\citep{vggt} and concurrent methods~\citep{wang2025pi,lin2025depth,keetha2026mapanything,vggt_omega} process all views jointly through a DINOv2~\citep{oquab2023dinov2}-based transformer, with extensions to incremental capture~\citep{wang2025spann3r,cut3r}, large view counts~\citep{yang2025fast3R, vggttt}, pose-free Gaussian splatting~\citep{ye2025noposplat}, multiple dense geometric quantities~\citep{dens3r} and dynamic scenes~\citep{zhang2025monst3r, vdpm, luo20264rc}. The joint multi-view baselines we evaluate use independently parameterized transformer layers with a fixed inference depth. We instead frame multi-view reconstruction as iterative refinement, matching the quality of these deeper feed-forward networks at a fraction of the parameter count while exposing the number of refinement steps as an inference-time compute knob.

\parsection{Iterative refinement.}
RAFT~\citep{RAFT} introduced GRU-based iterative refinement of a per-pixel flow field, an idiom since broadly applied to stereo matching~\citep{lipson2021raft}, visual SLAM~\citep{DroidSLAM,huang2025vipe}, and multi-view stereo~\citep{wang2022itermvs}: in each case, a lightweight recurrent updater iteratively refines geometry on top of a fixed feature backbone. iLRM~\citep{kang2026ilrm} extends iterative refinement to feed-forward 3D Gaussian splatting by treating successive (unshared) transformer layers as optimization steps over a scene representation decoupled from the input views. 
Unlike RAFT-style designs, where a lightweight recurrent updater iterates on a precomputed correspondence volume from a fixed feature backbone, our recurrence interleaves cross-view reasoning and refinement: each step is a full transformer block with frame and global attention, applied to per-view tokens initialized from a DINOv2 patch encoder. Cross-view reasoning and refinement therefore occur within the same repeated computation, rather than being assigned to separate stages.
Unlike iLRM, the block is shared across all $K$ steps and refines the per-view tokens themselves rather than a decoupled scene state.

LightGlue~\citep{lindenberger2023lightglue} also varies inference compute for image matching, adapting computation to the difficulty of an image pair. Our model instead exposes a user-selected reconstruction budget: we choose the number of applications of a shared block and recompute its time schedule, rather than choosing when to halt based on matching confidence.

\parsection{Weight-tied transformers.}
Applying a shared transformer block repeatedly across depth dates back to Universal Transformers~\citep{dehghani2019universal}, which paired weight tying with adaptive halting~\citep{graves2016adaptive}, and ALBERT~\citep{lan2020albert}, which showed that cross-layer parameter sharing yields competitive language models at a fraction of the parameters. Follow-ups have explored weight-tied recurrence in language modeling~\citep{hutchins2022blockrecurrent,yang2024looped,saunshi2025reasoning,geiping2025scaling} and learned-iteration in algorithmic tasks where networks trained for $K$ steps extrapolate to $K' > K$ at test time~\citep{schwarzschild2021algorithm,bansal2022endtoend}.
RAPTOR~\citep{jacobs2025raptor} showed that trained Vision Transformers admit an analogous structure: the layers can be faithfully approximated by a much smaller set of looped blocks, fit via post-hoc distillation against the original network. We adopt RAPTOR's gated block design but apply it differently. Rather than distilling a pretrained network into a looped form, we train a single shared block end-to-end on a 3D reconstruction task loss, with no teacher network and no distillation targets, and surpass an otherwise identical decoupled-parameters variant (\cref{tab:diag_block}) under matched training data and compute.

\section{Method}
\label{sec:method}

\subsection{Problem setup}
\label{sec:method_problem}
Given a set of $V$ input images $\{\im_i\}_{i=1}^{V}$ with $\im_i \in \real^{H \times W \times 3}$, our goal is to recover the underlying 3D scene geometry, expressed in the coordinate frame of the first view. We adopt a depth-ray representation: for each image $\im_i$, the model predicts a per-pixel depth map $\depthmap_i \in \real^{H \times W}$ and a dense ray map $\raymap_i \in \real^{H \times W \times 6}$.
Each pixel of the ray map encodes a 3D origin $\raymap^o \in \real^3$ and an unnormalized direction $\raymap^d \in \real^3$, so that a 3D point in world coordinates is obtained as $\mathbf{X} = \raymap^o + \depthmap(u,v) \cdot \raymap^d$.
Per-view camera-to-world rotations $\mathbf{R}_i \in SO(3)$, translations $\mathbf{t}_i \in \real^3$, and intrinsic matrices $\mathbf{K}_i \in \real^{3 \times 3}$ are recovered from the predicted ray maps following~\citet{lin2025depth}.

\subsection{Hypothesis}
\label{sec:hypothesis}

Two recent analyses motivate our approach.
First, \citet{jacobs2025raptor} show via post-hoc distillation that the $L$ layers of a trained ViT can be accurately approximated by a few looped blocks.
Second, \citet{stary2025understanding} probe DUSt3R~\citep{dust3r} layer by layer and show that its predicted pointmaps progressively refine across decoder depth, revealing iterative refinement of geometry inside multi-view transformers even though their layers do not share weights.

We hypothesize that this structure can be made explicit by applying a looped block to an evolving state, and that the resulting recurrence performs \emph{directional refinement} of the state, where the direction of $\z_k$ converges to the endpoint direction over the trained step range
(\cref{sec:method_recurrence,sec:analysis}).
In contrast to task-space iterative refinement methods such as RAFT~\citep{RAFT,lipson2021raft}, which decode after every step and apply a sequence loss on the output, we refine an internal state with a looped block and supervise only at the final step. This avoids running the decoder and computing a loss at every intermediate step, sparing $(K{-}1)$ decoder forward and backward passes per training iteration.

We model the recurrence as a time-conditioned discrete update over the partition $0 = t_0 < t_1 < \cdots < t_K = 1$ of the unit interval:
\begin{equation}
    \z_{k+1} = f_\theta(\z_k,\, t_k,\, t_{k+1})\,,
\label{eq:discrete_recurrence}
\end{equation}
where $f_\theta$ is a looped block conditioned on the continuous time interval $(t_k, t_{k+1})$.
Conditioning on continuous time, rather than on a discrete step index as in prior weight-tied transformers~\citep{dehghani2019universal}, decouples the block from any specific value of $K$ and lets a single set of weights cover a range of step counts at inference.
\Cref{sec:method_recurrence} instantiates $f_\theta$ as a shared transformer block with frame and global attention.

\begin{figure*}[t]
  \centering
  \includegraphics[width=\linewidth]{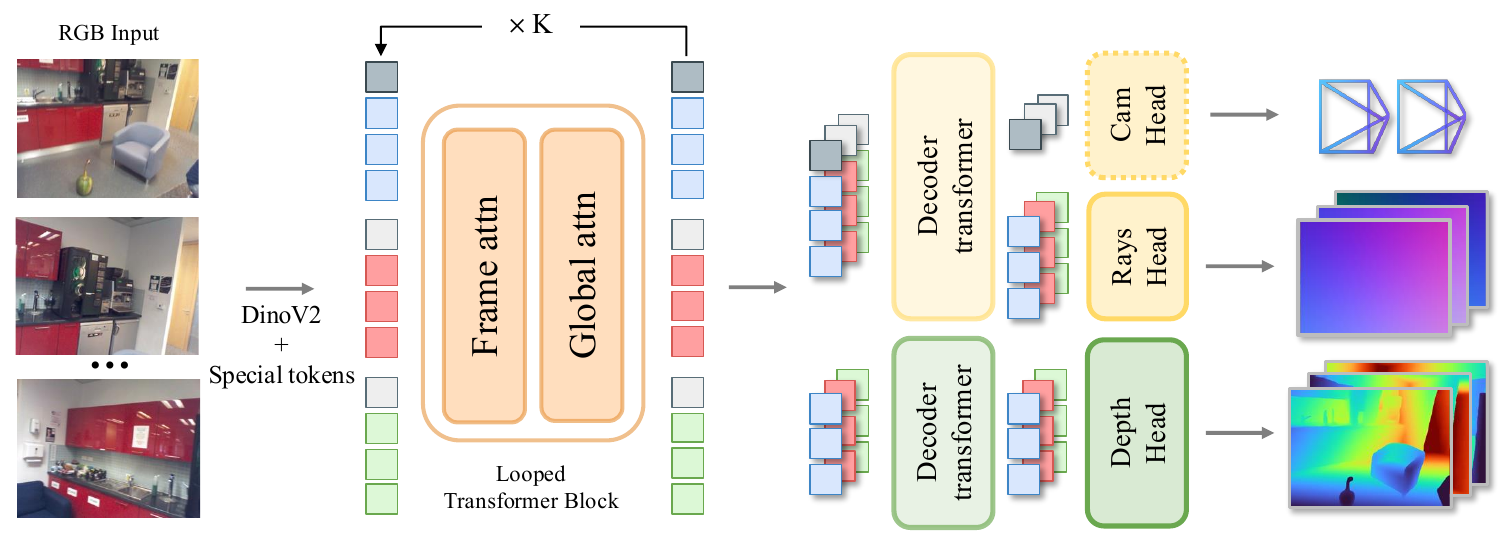}
    \caption{\textbf{Method overview.} $V$ input images are encoded by a shared DINOv2~\citep{oquab2023dinov2} backbone. A single looped transformer block with frame-wise and global attention sub-blocks is then applied recurrently to the resulting tokens for $K$ steps, with $K$ sampled per batch from $[K_\text{min}, K_\text{max}]$ during training. Two heads decode the final tokens into per-view depth and ray predictions.}
  \label{fig:method_overview}
  \vspace{-4mm}
\end{figure*}

\subsection{Architecture}
\label{sec:method_arch}

\parsection{Patch encoder and tokens.}
We initialize the per-view state $\z_0$ from a pretrained DINOv2~\citep{oquab2023dinov2} encoder, which maps each input image to an $\frac{H}{P} \times \frac{W}{P}$ grid of patch tokens.
We prepend a per-view copy of $R$ learnable register tokens~\citep{darcet2024vision} and of a learnable camera token to each view's token sequence, with the underlying parameters tied across views. The camera token uses two parameter sets: one for the reference (first) view, and one tied across all other views.
We encode patch positions with 2D rotary position embeddings~\citep{heo2024ropevit} and assign special tokens a sentinel position outside the patch grid.
We then apply a looped transformer block $K$ times to $\z_0$, with $K$ randomly sampled per training batch (\cref{sec:method_recurrence}), yielding the final state $\z_K$ passed to the decoder heads.

\parsection{Decoder heads.}
We pass $\z_K$ through two parallel decoder branches, each comprising a shallow transformer followed by an output head.
Each decoder transformer uses the same pre-norm $\text{Attn}+\text{MLP}$ block design as the shared recurrent block (\cref{sec:method_recurrence}), but without LayerScale.
The ray decoder uses a linear pixel-shuffle head to produce the per-pixel ray map $\raymap_\theta \in \real^{H \times W \times 6}$.
The depth decoder uses a convolutional depth head following~\citet{wang2025moge} to avoid block artifacts at patch boundaries (\cref{sec:suppl_two_stage}), and produces the depth map $\depthmap_\theta \in \real^{H \times W}$ and a depth confidence map $c_\depthmap \in \real^{H \times W}$.
Following \citet{lin2025depth}, we additionally include a camera MLP head that decodes a tuple $\mathbf{c}_\theta = (\mathbf{t}_\theta, \mathbf{q}_\theta, \mathbf{f}_\theta) \in \real^3 \times \mathbb{S}^3 \times \real^2$, comprising translation, unit rotation quaternion, and field of view, from the per-view camera tokens at the output of the ray decoder (\cref{sec:method_training}). It provides a faster alternative to the rays-derived recovery of \cref{sec:method_problem}, which remains our default at inference.
We obtain world points analytically as $\mathbf{X}_\theta = \raymap_\theta^o + \depthmap_\theta \cdot \raymap_\theta^d$.

\begin{figure}[t]
    \centering
    \begin{subfigure}[t]{\linewidth}
        \centering
        \includegraphics[width=0.82\linewidth]{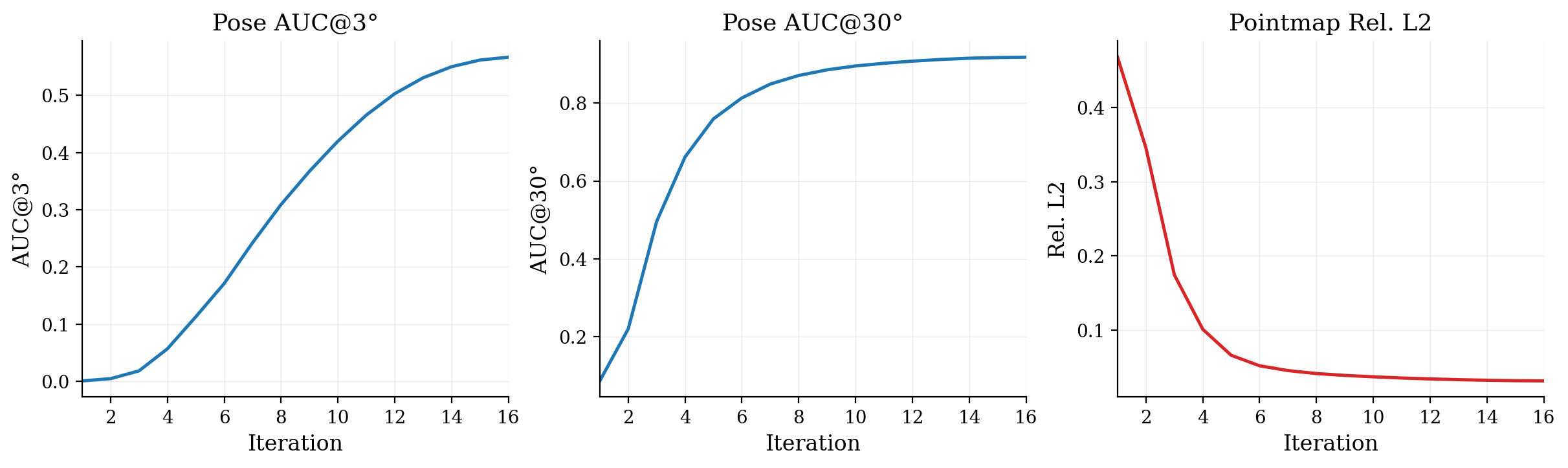}
        \caption{%
            \textbf{Task quality across recurrent iterations.}
            Decoding the residual stream $\z_k$ at every iteration $k\!\in\!\{1,\dots,16\}$ yields monotone improvement of pose and pointmap metrics across the trained step range.%
        }
        \label{fig:refinement-metrics}
    \end{subfigure}
    \begin{subfigure}[t]{\linewidth}
        \centering
        \includegraphics[width=0.82\linewidth]{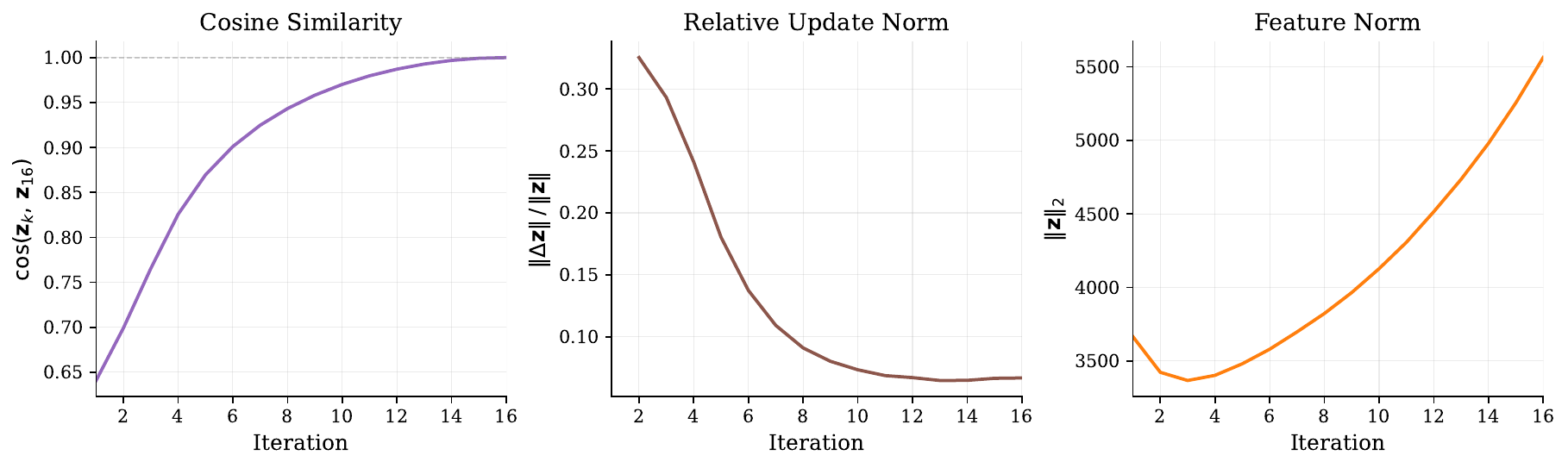}
        \caption{%
            \textbf{Residual-stream convergence.}
            \emph{Left:} cosine similarity between $\z_k$ and the final $\z_{16}$. \emph{Middle:} relative update norm $\lVert \Delta \z_k \rVert / \lVert \z_k \rVert$. \emph{Right:} feature norm $\lVert \z_k \rVert_{2}$.%
        }
        \label{fig:refinement-convergence}
    \end{subfigure}
    \caption{%
        \textbf{Iterative refinement of the residual stream.}
        Per-iteration analysis of \methodname's recurrent block, averaged across the five benchmarks of \cref{tab:comparison_sota_pmap,tab:comparison_sota_pose}.
        Task quality improves monotonically with the iteration count \textbf{(top)}. The recurrence does not contract to a fixed point in feature space (the state norm grows) but its direction stabilizes, with cosine similarity to the final state approaching $1$ and the relative update norm decaying by roughly $5{\times}$ \textbf{(bottom)}. These are empirical signatures over the observed rollout, rather than a guarantee of convergence beyond the trained range.%
    }
    \label{fig:refinement}
    \vspace{-4mm}
\end{figure}

\subsection{Looped Block}
\label{sec:method_recurrence}

\parsection{Block design.}
The looped block consists of two attention sub-blocks applied in sequence, following the alternating frame/global design of VGGT~\citep{vggt}.
The first is a frame attention that processes each view independently with 2D rotary position embeddings.
The second is a global attention that operates over the joint sequence of all tokens across all views.
Each sub-block uses a standard pre-norm $\text{Attn}+\text{MLP}$ design with LayerScale~\citep{touvron2021going}.

We condition the block on the time interval $(t_k, t_{k+1})$.
Three channel-wise scale vectors $(\mathbf{s}_\text{attn}, \mathbf{s}_\text{mlp}, \mathbf{s}_\text{out})$ control the block update:
\begin{equation}
\begin{aligned}
    & \z' && = \z_k + \mathbf{s}_\text{attn} \odot \text{LS}_1(\text{Attn}(\text{LN}_1(\z_k)))\,, \\
    & \z'' && = \z' + \mathbf{s}_\text{mlp} \odot \text{LS}_2(\text{MLP}(\text{LN}_2(\z')))\,, \\
    & \z_{k+1} && = \mathbf{s}_\text{out} \odot \z''\,,
\end{aligned}
\label{eq:gated_block}
\end{equation}
where $\text{LN}$ is layer normalization, $\text{LS}$ is LayerScale, and $\odot$ is channel-wise multiplication broadcast over the sequence dimension.
We compute the scales via a zero-initialized MLP such that $\mathbf{s} = \mathbf{1} + \text{MLP}(\gamma(t_k, t_{k+1}))$, where $\gamma$ concatenates the sinusoidal embeddings of $t_k$ and $t_{k+1}$.
We ablate simpler variants of this block (no $\mathbf{s}_\text{out}$, no time conditioning, and untied per-step blocks) in \cref{sec:analysis}.

\parsection{Variable step count.}
We train a single set of weights to serve as a 
$K$-elastic
family: the same block supports a range of step counts $K$ at inference, exposed as a compute knob.
Specifically, we sample $K \sim \text{Beta}(\alpha, \beta)$ per batch, scaled and rounded into $[K_\text{min}, K_\text{max}]$, and apply the block $K$ times along the uniform partition $0 = t_0 < t_1 < \cdots < t_K = 1$ with $t_k = k/K$,
where the application from $\z_k$ to $\z_{k+1}$ is conditioned on the interval $(t_k, t_{k+1})$.
At inference, we choose $K_\text{inf}$ and recompute the uniform grid $t_k=k/K_\text{inf}$, then execute the full schedule; varying $K_\text{inf}$ trades compute for accuracy within the trained range $[K_\text{min}, K_\text{max}]$, with degradation observed when $K_\text{inf}$ is pushed substantially outside it (\cref{sec:analysis}, \cref{sec:beyond_kmax}).

\parsection{Directional refinement.}
Unlike deep equilibrium networks~\citep{bai2019deep}, our recurrence does not converge to a fixed point in feature space. Instead, after a short initial contraction, the state norm $\|\z_k\|$ continues to grow over the observed rollout.
However, two empirical signatures characterize its dynamics within the trained step range (\cref{fig:refinement}).
First, $\cos(\z_k, \z_K)$ rises monotonically toward $1$ as $k \to K$, which means that each step moves the state closer in direction to the endpoint.
Second, the relative update magnitude $\|\Delta \z_k\|/\|\z_k\|$ decays from $\sim 0.5$ at the first step to $\sim 0.1$ at the last, indicating smaller updates relative to the current state norm.
We refer to this observed combination of increasing directional alignment and decreasing relative updates as \emph{directional refinement}. The improvement in decoded geometry is measured separately in \cref{fig:refinement}; these finite-rollout observations do not establish fixed-point convergence or guarantee improvement beyond the trained range.

\subsection{Training}
\label{sec:method_training}

\parsection{Losses.}
We supervise the model directly on the predicted geometry with five complementary
loss terms
covering depth, rays, world points, and camera parameters.
Following DUSt3R~\citep{dust3r}, predictions and ground truth are independently normalized prior to loss computation: given valid 3D points $\{\mathbf{X}_j\}_{j \in \Omega}$, we define the inverse normalization scale
\begin{equation}
    s = \left(\frac{1}{|\Omega|} \sum_{j \in \Omega} \lVert \mathbf{X}_j \rVert_2\right)^{\!-1}\,,
\label{eq:normalization_scale}
\end{equation}
computed separately for the predicted ($\hat{s}$) and ground-truth ($\bar{s}$) point clouds.
The per-sample training loss is then:
\begin{equation}
\begin{split}
    \loss ={} & \lVert \hat{s}\,\depthmap_\theta - \bar{s}\,\depthmap \rVert_2 + \loss_\text{grad}(\hat{s}\,\depthmap_\theta, \bar{s}\,\depthmap) + \lVert \hat{s}\,\raymap_\theta - \bar{s}\,\raymap \rVert_1 \\
    & + \lVert \hat{s}\,\mathbf{X}_\theta - \bar{s}\,\mathbf{X} \rVert_2 + \loss_\text{cam}(\mathbf{c}_\theta, \mathbf{c})\,,
\end{split}
\label{eq:total_loss}
\end{equation}
where $\mathbf{X}_\theta = \raymap_\theta^o + \depthmap_\theta \cdot \raymap_\theta^d$ is the analytically derived predicted point cloud, $\loss_\text{grad}$ is a multi-scale $\ell_1$ loss on horizontal and vertical depth gradients~\citep{Hu2019RevisitingSI, lin2025depth}, and $\loss_\text{cam}$ decomposes into separately weighted $\ell_1$ terms on the translation, rotation, and field-of-view components of $\mathbf{c}_\theta$.
The pointmap term ties the depth and ray heads through a joint geometric signal. %

\begin{table}[t]
    \caption{%
        \textbf{Pointmap accuracy.} Relative $\ell_2$ distance (\textbf{Rel.~L2}~$\downarrow$) and inlier ratio (\textbf{IR}~$\uparrow$) on the global pointmap after a Sim(3) alignment to the ground truth, across five benchmarks. The \colorbox{tabred}{best}, \colorbox{taborange}{second}, and \colorbox{tabyellow}{third} ranked results are highlighted. VGGT-$\Omega$~\citep{vggt_omega} is concurrent work, reported here for completeness.
    }
    \label{tab:comparison_sota_pmap}
    \footnotesize
    \setlength{\tabcolsep}{4pt}
    \centering
    \noindent\makebox[0.85\linewidth][c]{%
    \begin{tabular}{l rr rr rr rr rr}
        \toprule
        \multirow{2}{*}[-4pt]{\textbf{Method}}
            & \multicolumn{2}{c}{\textbf{DTU}} & \multicolumn{2}{c}{\textbf{ETH3D}} & \multicolumn{2}{c}{\textbf{7-Scenes}} & \multicolumn{2}{c}{\textbf{ScanNet++}} & \multicolumn{2}{c}{\textbf{nuScenes}} \\
        \cmidrule(lr){2-3}\cmidrule(lr){4-5}\cmidrule(lr){6-7}\cmidrule(lr){8-9}\cmidrule(lr){10-11}
            & Rel.~L2 & IR & Rel.~L2 & IR & Rel.~L2 & IR & Rel.~L2 & IR & Rel.~L2 & IR \\
        \midrule
        MASt3R~\citep{leroy2025grounding}                       & \cellcolor{tabyellow}0.011 & 94.9 & 0.340 & 29.1 & 0.076 & 41.7 & 0.251 & 14.5 & 0.360 & 11.4 \\
        MASt3R-SfM~\citep{duisterhof2024mast3rsfm}              & \cellcolor{tabred}0.009 & 96.9 & 0.095 & 54.1 & 0.051 & 64.7 & 0.042 & 69.7 & 0.311 & 18.4 \\
        MapAnything~\citep{keetha2026mapanything}               & 0.014 & 95.2 & 0.227 & 40.6 & 0.044 & 67.9 & \cellcolor{tabyellow}0.019 & \cellcolor{tabyellow}89.2 & 0.089 & \cellcolor{tabyellow}51.9 \\
        Pi3~\citep{wang2025pi}                                  & \cellcolor{tabred}0.009 & \cellcolor{tabred}97.3 & \cellcolor{tabyellow}0.034 & \cellcolor{tabyellow}66.8 & \cellcolor{tabred}0.032 & \cellcolor{tabred}77.8 & \cellcolor{tabred}0.014 & \cellcolor{tabred}94.3 & \cellcolor{tabyellow}0.078 & 51.0 \\
        VGGT~\citep{vggt}                                       & \cellcolor{taborange}0.010 & 95.8 & 0.053 & 52.6 & 0.042 & 70.8 & 0.034 & 68.4 & 0.081 & 42.3 \\
        VGGT-$\Omega$-1B~\citep{vggt_omega}                        & \cellcolor{tabred}0.009 & \cellcolor{taborange}97.2 & \cellcolor{tabred}0.024 & \cellcolor{tabred}78.6 & 0.039 & 65.3 & 0.032 & 70.6 & \cellcolor{tabred}0.055 & \cellcolor{tabred}62.3 \\
        DA3-L~\citep{lin2025depth}                              & \cellcolor{taborange}0.010 & \cellcolor{tabyellow}97.1 & 0.211 & 49.9 & 0.039 & 69.6 & 0.051 & 48.1 & 0.141 & 27.0 \\
        DA3-G~\citep{lin2025depth}                              & \cellcolor{taborange}0.010 & \cellcolor{tabyellow}97.1 & 0.129 & 64.7 & \cellcolor{tabyellow}0.037 & \cellcolor{tabyellow}71.1 & 0.041 & 57.8 & 0.080 & 42.0 \\
        \midrule
        \textbf{Ours}                                           & \cellcolor{tabred}0.009 & \cellcolor{tabyellow}97.1 & \cellcolor{taborange}0.026 & \cellcolor{taborange}78.3 & \cellcolor{taborange}0.035 & \cellcolor{taborange}74.2 & \cellcolor{taborange}0.015 & \cellcolor{taborange}93.3 & \cellcolor{taborange}0.067 & \cellcolor{taborange}58.5 \\
        \bottomrule
    \end{tabular}}
\end{table}

\begin{table}[t]
    \caption{%
        \textbf{Camera pose accuracy.} Area under the cumulative pose-error curve at $3^{\circ}$ (\textbf{AUC@3}~$\uparrow$) and $30^{\circ}$ (\textbf{AUC@30}~$\uparrow$), across five benchmarks.         VGGT-$\Omega$~\citep{vggt_omega} is concurrent work, reported here for completeness. \methodname{} ranks first or second on nine of ten cells and is in the top three on every cell.
    }
    \label{tab:comparison_sota_pose}
    \footnotesize
    \setlength{\tabcolsep}{6pt}
    \centering
    \noindent\makebox[0.85\linewidth][c]{%
    \begin{tabular}{l rr rr rr rr rr}
        \toprule
        \multirow{2}{*}[-4pt]{\textbf{Method}}
            & \multicolumn{2}{c}{\textbf{DTU}} & \multicolumn{2}{c}{\textbf{ETH3D}} & \multicolumn{2}{c}{\textbf{7-Scenes}} & \multicolumn{2}{c}{\textbf{ScanNet++}} & \multicolumn{2}{c}{\textbf{nuScenes}} \\
        \cmidrule(lr){2-3}\cmidrule(lr){4-5}\cmidrule(lr){6-7}\cmidrule(lr){8-9}\cmidrule(lr){10-11}
        \multicolumn{1}{r}{\normalfont AUC}
            & @3 & @30 & @3 & @30 & @3 & @30 & @3 & @30 & @3 & @30 \\
        \midrule
        MASt3R~\citep{leroy2025grounding}                       & 21.6 & 81.7 & 35.2 & 57.6 & 9.3 & 70.6 & 15.5 & 43.9 & 9.4 & 62.7 \\
        MASt3R-SfM~\citep{duisterhof2024mast3rsfm}              & 40.2 & 91.4 & 52.8 & \cellcolor{taborange}85.0 & \cellcolor{taborange}16.4 & 80.3 & 38.7 & 85.0 & 9.1 & 70.9 \\
        MapAnything~\citep{keetha2026mapanything}               & 18.1 & 88.8 & 37.1 & 73.8 & 8.2 & 74.6 & \cellcolor{tabyellow}70.6 & \cellcolor{tabyellow}96.7 & 37.8 & 82.8 \\
        Pi3~\citep{wang2025pi}                                  & 70.2 & 97.3 & 43.2 & \cellcolor{tabyellow}84.9 & 11.3 & \cellcolor{tabyellow}81.3 & \cellcolor{taborange}76.5 & \cellcolor{taborange}97.3 & 24.8 & 82.4 \\
        VGGT~\citep{vggt}                                       & \cellcolor{tabred}96.5 & \cellcolor{tabred}99.8 & 35.4 & 82.8 & 11.1 & 79.1 & 15.1 & 74.7 & \cellcolor{tabyellow}39.2 & \cellcolor{tabyellow}83.8 \\
        VGGT-$\Omega$-1B~\citep{vggt_omega}                        & \cellcolor{tabyellow}77.4 & \cellcolor{tabyellow}98.1 & \cellcolor{taborange}64.1 & \cellcolor{tabred}95.4 & \cellcolor{tabred}21.3 & \cellcolor{tabred}87.0 & 29.9 & 87.3 & \cellcolor{taborange}42.4 & \cellcolor{tabred}85.5 \\
        DA3-L~\citep{lin2025depth}                              & 69.2 & 97.3 & 38.8 & 75.3 & 11.8 & 79.7 & 7.6 & 71.8 & 11.0 & 74.7 \\
        DA3-G~\citep{lin2025depth}                              & 74.9 & 97.9 & \cellcolor{tabyellow}55.4 & 82.4 & 12.0 & 80.9 & 26.4 & 80.2 & 37.7 & 82.0 \\
        \midrule
        \textbf{Ours}                                           & \cellcolor{taborange}83.2 & \cellcolor{taborange}98.8 & \cellcolor{tabred}66.0 & \cellcolor{tabred}95.4 & \cellcolor{tabyellow}13.9 & \cellcolor{taborange}81.7 & \cellcolor{tabred}79.4 & \cellcolor{tabred}98.0 & \cellcolor{tabred}43.4 & \cellcolor{taborange}85.3 \\
        \bottomrule
    \end{tabular}}
\end{table}

\parsection{Optimization.}
Training proceeds in two stages.
The first stage trains the model end-to-end with all loss terms, using plain $\ell_2$ regression on the depth and pointmap terms and a linear pixel-shuffle depth head.
The second stage swaps in the convolutional depth head described in \cref{sec:method_arch} and finetunes the depth decoder while freezing all other parameters. The ray and camera losses are disabled in the second stage. The depth term applies DUSt3R-style confidence weighting~\citep{dust3r}, $\loss_\text{unc}(c, \mathbf{a}, \mathbf{b}) = c \lVert \mathbf{a} - \mathbf{b} \rVert_2 - \lambda_c \log c$, parameterized by the predicted per-pixel uncertainty $c_\depthmap$, while the pointmap term remains plain $\ell_2$ on the analytically derived points.

\section{Experiments}
\label{sec:experiments}

\parsection{Implementation.}
We implement our method in PyTorch~\citep{paszke2019pytorch} and train on 128 H100 GPUs with $V \in [2, 18]$ views per scene at 504-pixel longest-edge resolution. Each training step uses up to $4{,}608$ images and a fixed tokens budget ($\approx$2.5M tokens).
We use a DINOv2 ViT-B encoder with embedding dimension $768$, patch size $P{=}14$, and $R{=}4$ register tokens. The ray and depth decoders each use embedding dimension $384$ and two transformer blocks.
We sample $K \sim \text{Beta}(2, 1)$ scaled into $[8, 16]$ during training so that a single checkpoint supports any step count in this range (\cref{tab:diag_iterative}). We run $K_\text{inf} = 16$ steps at inference.
We optimize with AdamW~\citep{loshchilov2018decoupled} at a base learning rate of $3 \times 10^{-4}$, weight decay $0.05$, and a cosine decay schedule without warmup in the first stage, applying a $0.1\times$ multiplier to the DINOv2 backbone.
The first stage trains end-to-end for 200K iterations. The second stage finetunes the depth decoder (see \cref{sec:method_training}) for 40K iterations at $1 \times 10^{-4}$ with a 500-step linear warmup and a confidence regularizer weight of $\lambda_c = 0.2$.
The depth, ray, pointmap, and camera losses are weighted equally, and within the camera loss, the translation, rotation, and field-of-view terms are weighted $(1, 1, 0.5)$. 
We apply the depth-gradient term only on synthetic data, where ground-truth depth is dense enough for reliable gradient supervision. We train on a mixture of 29 public datasets and list them in the supplemental material.

\subsection{Comparison with State of the Art}
\label{sec:sota}

\parsection{Evaluation datasets.}
We evaluate on a set of diverse benchmarks that span indoor, outdoor, object-centric, and driving scenes:
DTU~\citep{jensen2014dtu},
ETH3D~\citep{schops2017eth3d},
7-Scenes~\citep{shotton2013scene},
ScanNet++~\citep{yeshwanth2023scannetpp},
and nuScenes~\citep{caesar2020nuscenes}. For ScanNet++, which serves as a training dataset for multiple baselines~\citep{lin2025depth, wang2025pi, keetha2026mapanything, duisterhof2024mast3rsfm, leroy2025grounding, vggt_omega} and our model, we enforce a clean scene-level split between the training and evaluation data.

\parsection{Evaluation metrics.}
We evaluate reconstruction quality with two metrics computed on the global pointmap after a Sim(3) alignment of the predicted points to the ground truth. Both are derived from the per-point relative error $r_i = \lVert \mathbf{X}_{\theta,i} - \mathbf{X}_{\text{gt},i}\rVert / \lVert \mathbf{X}_{\text{gt},i}\rVert$: the relative $\ell_2$ distance (\textbf{Rel.~L2}~$\downarrow$) is its mean over valid points, and the inlier ratio (\textbf{IR}~$\uparrow$) is the fraction of points with $r_i < 3\%$.
For camera pose accuracy, we report the area under the cumulative error curve at angular thresholds of $3^{\circ}$ and $30^{\circ}$ (\textbf{AUC@3}~$\uparrow$ and \textbf{AUC@30}~$\uparrow$), where the per-pair error is the maximum of the rotation and translation angle errors.

\parsection{Baselines.}
We compare against state-of-the-art feed-forward 3D reconstruction methods: VGGT~\citep{vggt}, Pi3~\citep{wang2025pi}, MapAnything~\citep{keetha2026mapanything}, and Depth Anything 3 (DA3)~\citep{lin2025depth}. We also include MASt3R~\citep{leroy2025grounding} and MASt3R-SfM~\citep{duisterhof2024mast3rsfm}, which combine pairwise prediction with sparse global alignment. In addition, we report numbers for the concurrent work VGGT-$\Omega$~\citep{vggt_omega} for completeness. We run all baselines through our evaluation framework using their official released checkpoints, with MapAnything and DA3 at their v1.1 releases, VGGT-$\Omega$ at the 1B-512 release, and DA3 reported at two backbone scales (ViT-L and ViT-G). Per-baseline configurations are detailed in \cref{sec:suppl_baselines}.
These comparisons measure the performance of released models under our evaluation protocol. Their pretraining, dataset mixtures, and training budgets differ, so performance differences cannot be attributed to architecture alone. We isolate the effect of weight sharing within our own architecture in \cref{sec:analysis}.

\begin{table}[t]
    \caption{
        \textbf{Model efficiency and quality.}
        Parameter count, forward-pass FLOPs (total and per-image), peak inference GPU memory, and average IR / AUC@$30^{\circ}$ across the five benchmarks of \cref{tab:comparison_sota_pmap,tab:comparison_sota_pose}, measured on a single A100 with $24$ input views.
        For MASt3R~\citep{leroy2025grounding} (\texttt{swin-5}, 120 pairs) and MASt3R-SfM~\citep{duisterhof2024mast3rsfm} (retrieval-20-10, 273 pairs), the reported FLOPs cover only the pair-network forward passes, excluding the iterative global alignment that follows. Their lower peak memory comes from processing pairs sequentially.
        \methodname{} leads on average IR and AUC@$30^{\circ}$ at the smallest parameter count.
    }
    \label{tab:comparison_efficiency}
    \footnotesize
    \setlength{\tabcolsep}{5pt}

    \centering
    \noindent\makebox[0.85\linewidth][c]{%
    \begin{tabular}{l rrrr|rr}
        \toprule
        \multirow{2}{*}{\textbf{Method}}
            & \textbf{Params} & \textbf{Compute} & \textbf{Compute / Img} & \textbf{Peak Mem}
            & \textbf{IR} & \textbf{AUC@30} \\
        & (M)~$\downarrow$ & (TFLOPs)~$\downarrow$ & (TFLOPs)~$\downarrow$ & (GiB)~$\downarrow$
        & (\%)~$\uparrow$ & (\%)~$\uparrow$ \\
        \midrule
        MASt3R~\citep{leroy2025grounding}                       & \cellcolor{tabyellow}689 & 500.0 & 20.8 & \cellcolor{taborange}4.4 & 38.3 & 63.3 \\
        MASt3R-SfM~\citep{duisterhof2024mast3rsfm}              & 690 & 1150.1 & 47.9 & \cellcolor{tabred}3.4 & 60.8 & 82.5 \\
        MapAnything~\citep{keetha2026mapanything}               & 1228 & 148.4 & 6.2 & 20.1 & 69.0 & 83.3 \\
        Pi3~\citep{wang2025pi}                                  & 959 & 153.8 & 6.4 & 6.6 & \cellcolor{taborange}77.4 & \cellcolor{tabyellow}88.6 \\
        VGGT~\citep{vggt}                                       & 1257 & 190.0 & 7.9 & 14.7 & 66.0 & 84.0 \\
        VGGT-$\Omega$-1B~\citep{vggt_omega}                        & 1144 & \cellcolor{tabyellow}99.8 & \cellcolor{tabyellow}4.2 & 7.9 & \cellcolor{tabyellow}74.8 & \cellcolor{taborange}90.7 \\
        DA3-L~\citep{lin2025depth}                              & \cellcolor{taborange}356 & \cellcolor{tabred}71.4 & \cellcolor{tabred}3.0 & 7.3 & 58.3 & 79.7 \\
        DA3-G~\citep{lin2025depth}                              & 1201 & 178.7 & 7.4 & 13.0 & 66.5 & 84.7 \\
        \midrule
        \textbf{Ours}                                           & \cellcolor{tabred}\textbf{117} & \cellcolor{taborange}\textbf{75.9} & \cellcolor{taborange}\textbf{3.2} & \cellcolor{tabyellow}\textbf{4.9} & \cellcolor{tabred}\textbf{80.3} & \cellcolor{tabred}\textbf{91.8} \\
        \bottomrule
    \end{tabular}}
    \vspace{-4mm}
\end{table}

\parsection{Results.}
At $117$M parameters and $75.9$ TFLOPs of compute (\cref{tab:comparison_efficiency}), \methodname{} matches or exceeds much larger feed-forward methods while running in under $5$ GiB of peak memory at $24$ input views.
Our model leads on average IR and AUC@$30^\circ$, although no method leads on every benchmark. Pi3 is stronger on indoor pointmap accuracy at roughly $8{\times}$ our parameters and $2{\times}$ our compute. VGGT leads on DTU pose, particularly at AUC@$3^\circ$ ($96.5$ vs $83.2$), while the gap is within one point at AUC@$30^\circ$. VGGT-$\Omega$ improves outdoor Rel.~L2 and 7-Scenes pose, but trails our average IR and AUC@$30^\circ$ at roughly $10{\times}$ our parameters. MASt3R-SfM is competitive indoors, with pair-network forward passes alone costing about $15{\times}$ our compute.

\parsection{Memory.}
Sharing weights reduces parameter storage, and our decoder consumes only the final recurrent state. It does not change the dense global-attention structure: as the number of views grows, activation and output storage become increasingly important. The memory figures in \cref{tab:comparison_efficiency} therefore characterize the measured 24-view setting, rather than an asymptotic memory advantage.

\begin{figure*}[t]
    \centering
    \includegraphics[width=\linewidth]{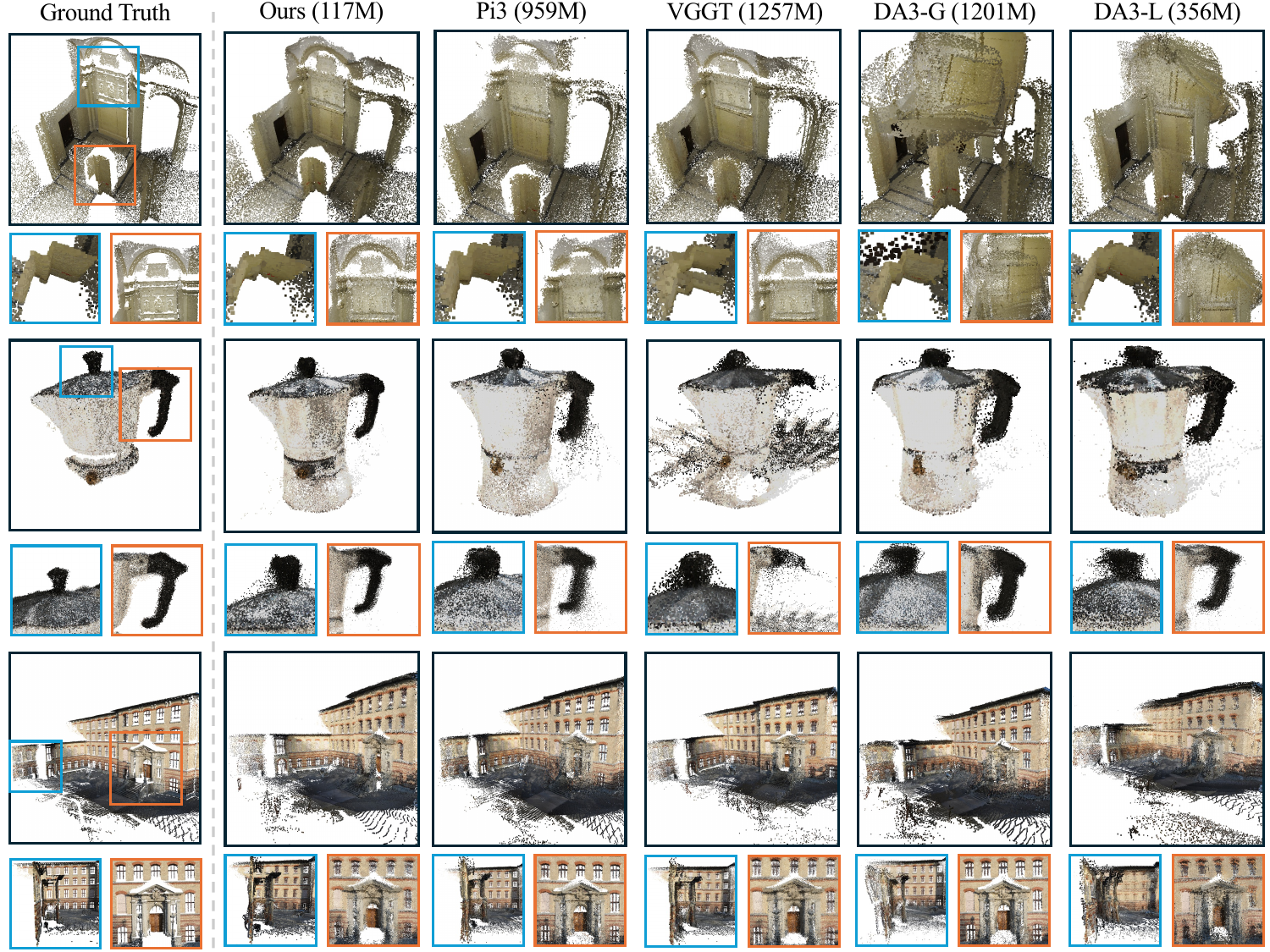}
    \caption{
        \textbf{Qualitative results.}
        Predicted point clouds for three scenes, comparing \methodname{} to four feed-forward baselines with parameter counts shown above each column. \methodname{} produces denser, less noisy point clouds despite using far fewer parameters.
    }
    \label{fig:qualitative}
    \vspace{-4mm}
\end{figure*}

\subsection{Analysis}
\label{sec:analysis}

We ablate block design (\cref{tab:diag_block}) and training/inference step count (\cref{tab:diag_iterative}).
All variants use a ViT-B encoder and are trained for 100K iterations on the same data, and differ only along the examined axis.

\parsection{Block design.}
We progressively add the components of our recurrent design in \cref{tab:diag_block}, starting from a fully decoupled variant in which each of the 16 recurrent steps has its own block (no weight sharing, no time conditioning).
Weight sharing collapses the 16 independent blocks into a single shared block applied recurrently. The time-conditioned residual gates ($\mathbf{s}_\text{attn}, \mathbf{s}_\text{mlp}$) modulate the block's attention and MLP branches, and the state gate ($\mathbf{s}_\text{out}$) yields our full method.
Each component improves all four reported metrics. Weight sharing alone raises IR from $61.1$ to $66.4$ and AUC@$30^\circ$ from $82.0$ to $84.8$, while reducing the recurrent core's parameter count by $16{\times}$; the encoder and decoder parameters are unchanged. Adding the conditioned gates further raises these scores to $69.2$ and $86.9$.
\begin{table}[t]
    \caption{
        \textbf{Block design.}
        We progressively add the components of our looped block.
        Weight sharing constrains the recurrence to a single looped block, the time-conditioned residual gates modulate the attention and MLP branches, and the state gate modulates the residual stream.
        We report metrics averaged across the five benchmarks of \cref{tab:comparison_sota_pmap,tab:comparison_sota_pose}.
    }
    \label{tab:diag_block}
    \footnotesize
    \setlength{\tabcolsep}{6pt}
    \centering
    \noindent\makebox[0.85\linewidth][c]{%
    \begin{tabular}{l|ccc|rrrr}
        \toprule
        \multirow{2}{*}{\textbf{Variant}}
            & \makecell{Weight \\ sharing}
            & \makecell{Residual gates \\ ($\mathbf{s}_\text{attn}, \mathbf{s}_\text{mlp}$)}
            & \makecell{State gate \\ ($\mathbf{s}_\text{out}$)}
            & \multirow{2}{*}{\textbf{Rel.~L2} $\downarrow$}
            & \multirow{2}{*}{\textbf{IR} $\uparrow$}
            & \multirow{2}{*}{\textbf{AUC@3} $\uparrow$}
            & \multirow{2}{*}{\textbf{AUC@30} $\uparrow$} \\
        \midrule
        Decoupled & \xmark & \xmark & \xmark & 0.056 & 61.1 & 23.0 & 82.0 \\
        Shared & \cmark & \xmark & \xmark & \cellcolor{tabyellow}0.045 & \cellcolor{tabyellow}66.4 & \cellcolor{tabyellow}30.2 & \cellcolor{tabyellow}84.8 \\
        Shared + residual gates & \cmark & \cmark & \xmark & \cellcolor{taborange}0.042 & \cellcolor{taborange}67.0 & \cellcolor{taborange}31.5 & \cellcolor{taborange}85.9 \\
        \midrule
        \textbf{Shared + state gate} & \cmark & \cmark & \cmark & \cellcolor{tabred}0.040 & \cellcolor{tabred}69.2 & \cellcolor{tabred}33.3 & \cellcolor{tabred}86.9 \\
        \bottomrule
    \end{tabular}}
    \vspace{-4mm}
\end{table}

\parsection{Step-count.}
We vary the recurrent step count at training and inference in \cref{tab:diag_iterative}. The fixed-$K$ baselines train and run with a constant step count. Our variable-$K$ model trains with $K \sim \text{Beta}(2, 1)$ on $[8, 16]$ and is evaluated at $K_\text{inf}{=}12$ and $K_\text{inf}{=}16$.
At $K_\text{inf}{=}16$, variable-$K$ training stays within $0.5$ percentage points of the fixed-$16$ specialist on AUC@$3^\circ$; at $K_\text{inf}{=}12$, the gap to the fixed-$12$ specialist is $0.8$ points. Fixed models transfer less well across budgets: the fixed-$16$ model reaches only $24.5$ AUC@$3^\circ$ at 12 steps, compared with $29.6$ for variable-$K$, while running the fixed-$12$ model for 16 steps reduces its score from $30.4$ to $27.3$. Variable-$K$ training thus provides flexibility across budgets, rather than consistently improving on a specialist at its own budget.

Using a shorter schedule also differs from early-stopping a longer rollout. With the full model at $K_\text{inf}{=}8$, recomputing the time grid raises AUC@$3^\circ$ from $0.31$ to $0.44$ compared with truncating a 16-step schedule (\cref{fig:kinf_intermediate}).

\begin{table}[!t]
    \caption{
        \textbf{Step-count analysis.}
        Fixed-$K$ baselines are evaluated both at their training budget and at the other budget. Our variable-$K$ model trains with $K \sim \text{Beta}(2, 1)$ on $[8, 16]$ and is evaluated at $K_\text{inf}{=}12$ and $K_\text{inf}{=}16$. Metrics are averaged across the five benchmarks of \cref{tab:comparison_sota_pmap,tab:comparison_sota_pose}.
        A single variable-$K$ checkpoint serves both budgets. $\dagger$ marks a step count not used to train the fixed-$K$ model.
    }
    \label{tab:diag_iterative}
    \footnotesize
    \setlength{\tabcolsep}{6pt}
    \centering
    \noindent\makebox[0.85\linewidth][c]{%
    \begin{tabular}{l|cc|rrrr}
        \toprule
        \multirow{2}{*}{\textbf{Variant}}
            & \multirow{2}{*}{\makecell{Training\\ $K$-sampler}}
            & \multirow{2}{*}{\makecell{$K_\text{inf}$}}
            & \multirow{2}{*}{\textbf{Rel.~L2} $\downarrow$}
            & \multirow{2}{*}{\textbf{IR} $\uparrow$}
            & \multirow{2}{*}{\textbf{AUC@3} $\uparrow$}
            & \multirow{2}{*}{\textbf{AUC@30} $\uparrow$} \\
            & & & & & & \\
        \midrule
            Fixed $K{=}12$ & fixed & 12 & 0.044 & 67.6 & 30.4 & 85.5 \\
            Fixed $K{=}16$ & fixed & $12^\dagger$ & 0.049 & 61.0 & 24.5 & 83.6 \\
            \textbf{Ours} (Variable, $K_\text{max}{=}16$) & Beta on $[8, 16]$ & 12 & 0.043 & 66.7 & 29.6 & 85.4 \\
        \midrule
            Fixed $K{=}12$ & fixed & $16^\dagger$ & 0.046 & 62.9 & 27.3 & 83.9 \\
            Fixed $K{=}16$ & fixed & 16 & 0.041 & 69.6 & 33.8 & 86.8 \\
            \textbf{Ours} (Variable, $K_\text{max}{=}16$) & Beta on $[8, 16]$ & 16 & 0.040 & 69.2 & 33.3 & 86.9 \\
        \bottomrule
    \end{tabular}}
    \vspace{-4mm}
\end{table}

\parsection{Correspondence refinement.}
We evaluate patch correspondence in our model's attention maps on MegaDepth-1500 and ScanNet-1500. Ground-truth patch correspondences are obtained by warping patch centers with depth and relative pose, retaining co-visible queries after an occlusion check. At each iteration, we restrict a query's post-softmax global-attention row to the target view's patch tokens and score its maximum against the ground-truth correspondence with a one-patch tolerance. \Cref{tab:correspondence_main} reports the same best-performing head across all iterations, alongside nearest-neighbor matching on the input DINOv2 features.
The measured hit rate increases across the 16 iterations, reaching $79.8\%$ on MegaDepth and $68.3\%$ on ScanNet, compared with $49.3\%$ and $40.2\%$ for DINOv2. Because the same block is reused at each step, this improvement shows increasingly accurate correspondence search with shared weights.
This suggests that the loop implements an iterative correspondence search, despite the model being supervised only on 3D reconstruction and pose estimation losses. We refer to \cref{sec:emergent_correspondence} for further details on the attention probing and visualizations.

\begin{table}[!htb]
\centering
\small
\caption{\textbf{Correspondence refinement.} Hit@1 (\%, one-patch tolerance) for the best-performing global-attention head at selected iterations. DINOv2 nearest-neighbor matching provides an input-feature reference. Both datasets have a reported chance reference of $0.9\%$.}
\label{tab:correspondence_main}
\begin{tabular}{l r rrrr}
\toprule
& DINOv2 NN & $k=1$ & $k=4$ & $k=8$ & $k=16$ \\
\midrule
MegaDepth-1500 & 49.3 & 44.6 & 71.2 & 78.4 & 79.8 \\
ScanNet-1500 & 40.2 & 42.1 & 62.5 & 67.2 & 68.3 \\
\bottomrule
\end{tabular}
\end{table}

\section{Discussion}
\label{sec:conclusion}

Modern 3D reconstruction transformers have improved primarily by scaling. We presented \methodname, which instead applies a single shared block recurrently to a DINOv2-initialized state, sampling the step count at training so that one checkpoint covers a range of inference budgets. It matches or surpasses much larger feed-forward baselines across five reconstruction benchmarks at a fraction of their parameters and comparable or lower compute, suggesting that parameter scaling is not the only path forward for 3D reconstruction.

\parsection{Limitations.}\label{sec:limitations}
The recurrence does not support arbitrary extrapolation beyond its trained step range: some feature channels grow without bound and performance eventually collapses (\cref{sec:beyond_kmax}). Preliminary input-injection variants inspired by~\citet{yang2024looped} did not prevent this degradation in our setting; stable extrapolation remains an open problem. Variable-$K$ training trades specialist quality for flexibility across budgets, and the model does not explicitly represent scene dynamics. 

\clearpage
\section*{Acknowledgements}
This project was partially funded by the ERC Starting Grant DynAI (ERC-101043189).
We thank Harsh Singh for his help with the experimental results during the rebuttal period.

\bibliographystyle{plainnat}
\bibliography{main}

\clearpage
\appendix
\FloatBarrier\begin{center}{\Large\textbf{Supplementary Material for Déjà View}}\end{center}
\section{Training and Evaluation Details}
We document the training mixture and baseline configurations used in the main evaluation.

\subsection{Training Datasets}
\label{sec:training_datasets}
We train on a mixture of 29 publicly available datasets that span synthetic
renderings, indoor and outdoor real captures, multi-view object scans, and
driving footage. The corpus is highly imbalanced: per-dataset image counts
$N_i$ span more than three orders of magnitude (from $\sim$10\,k for
Spring to $\sim$11.8\,M for Aria Synthetic Environments), so naive
proportional sampling would let a handful of large datasets dominate
training, while uniform sampling would massively oversample the smallest
ones. In LLM literature, multilingual training faces the same imbalance across high- and
low-resource languages, which is addressed with temperature sampling
\citep{arivazhagan2019massively}. We adopt
the same recipe, treating each dataset as a ``language'' with token
budget $N_i$: the probability of drawing a training
example from dataset $i$ is set to
\[
    p_i \;=\; \frac{N_i^{\alpha}}{\sum_j N_j^{\alpha}},
    \qquad \alpha = 0.5,
\]
i.e.\ proportional to $\sqrt{N_i}$, which flattens the head of the
distribution while still favouring the larger, more diverse corpora.
The realised epoch share for each dataset is reported in
Table~\ref{tab:training_datasets}.

\begin{table}[t]
    \centering
    \footnotesize
    \caption{Per-epoch training-mixture share for the 29 datasets, sorted
    by share. By construction, mix\,\% $= p_i \propto \sqrt{N_i}$ where
    $N_i$ is the total number of training images in dataset $i$.}
    \label{tab:training_datasets}
    \begingroup
\setlength{\tabcolsep}{3pt}
\begin{tabular}{@{}lr@{\hspace{0.8em}}lr@{}}
    \toprule
    \textbf{Dataset} & \textbf{mix \%} & \textbf{Dataset} & \textbf{mix \%} \\
    \midrule
    Aria Synth.\ Env.~\citep{avetisyan2024scenescript}                  & 13.77 & BEDLAM~\citep{black2023bedlam}                       & 1.99 \\
    WildRGB-D~\citep{xia2024wildrgbd}                                   &  9.32 & BlendedMVS~\citep{yao2020blendedmvs}                 & 1.35 \\
    TRELLIS~\citep{xiang2025trellis,deitke2023objaverse}                &  9.04 & MegaDepth~\citep{li2018megadepth}                    & 1.16 \\
    CO3D~\citep{reizenstein2021co3d}                                    &  8.84 & Replica~\citep{straub2019replica}                    & 0.95 \\
    Taskonomy~\citep{zamir2018taskonomy}                                &  7.26 & Virtual KITTI~2~\citep{cabon2020vkitti2}             & 0.83 \\
    ScanNet~\citep{dai2017scannet}                                      &  6.23 & Hypersim~\citep{roberts2021hypersim}                 & 0.69 \\
    DL3DV~\citep{ling2024dl3dv}                                         &  5.84 & Kubric~\citep{greff2022kubric}                       & 0.60 \\
    Cubify Any.~\citep{lazarow2025cubifyanything,baruch2021arkitscenes} &  5.73 & GTA-SfM~\citep{wang2020gtasfm}                       & 0.52 \\
    TartanAir~V2~\citep{wang2020tartanair}                              &  4.78 & MatrixCity~\citep{li2023matrixcity}                  & 0.49 \\
    Parallel Dom.\ 4D~\citep{vanhoorick2024gcd}                         &  4.75 & MPSD~\citep{lopez2020mpsd}                           & 0.49 \\
    ScanNet++~\citep{yeshwanth2023scannetpp}                            &  3.96 & Mapillary Metr.~\citep{mapillary2021metropolis}      & 0.47 \\
    Waymo~\citep{sun2020waymo}                                          &  2.76 & UnrealStereo4K~\citep{tosi2021smdnets}               & 0.44 \\
    Mid-Air~\citep{fonder2019midair}                                    &  2.61 & MVS-Synth~\citep{huang2018deepmvs}                   & 0.44 \\
    Dynamic Replica~\citep{karaev2023dynamicstereo}                     &  2.16 & Spring~\citep{mehl2023spring}                        & 0.40 \\
    TartanAir~\citep{wang2020tartanair}                                 &  2.13 &                                                      &      \\
    \bottomrule
\end{tabular}
\endgroup

\end{table}

\subsection{Baseline Configurations}
\label{sec:suppl_baselines}

We run all baselines through our evaluation framework using their official code releases and checkpoints, and apply the same Sim(3) alignment to predicted pointmaps before computing metrics (\cref{sec:sota}). For Rel.~L2 and IR we use each method's primary 3D output: depth-unprojected pointmaps for VGGT, DA3, and our model, the direct point-head output for Pi3 and MapAnything, and the SGA-optimized dense pointmap for MASt3R and MASt3R-SfM.

\paragraph{VGGT~\citep{vggt}.}
The official \texttt{facebook/VGGT-1B} checkpoint at $518$-pixel longest edge with patch size $14$.

\paragraph{VGGT-$\Omega$~\citep{vggt_omega}.}
The official \texttt{facebook/VGGT-Omega} 1B checkpoint (\texttt{vggt\_omega\_1b\_512.pt}, without text alignment) at $512$-pixel longest edge with patch size $16$. We decode the released 9D camera encoding (translation, quaternion, FoV$_h$, FoV$_w$) via the official \texttt{encoding\_to\_camera} utility and use depth-unprojected pointmaps as the primary 3D output, matching the convention used for VGGT and DA3.

\paragraph{Pi3~\citep{wang2025pi}.}
The official \texttt{yyfz233/Pi3} checkpoint at $518$-pixel longest edge with patch size $14$. %

\paragraph{MapAnything~\citep{keetha2026mapanything}.}
The official \texttt{facebook/map-anything} v1.1 checkpoint at $518$-pixel longest edge with patch size $14$.

\paragraph{Depth Anything 3 (DA3)~\citep{lin2025depth}.}
The official v1.1 checkpoints at two backbone scales (DA3-L: \texttt{depth-anything/DA3-LARGE}, ViT-L, $356$M params; DA3-G: \texttt{depth-anything/DA3-GIANT}, ViT-G, $1.2$B params), both at $504$-pixel longest edge with patch size $14$. We decode camera pose from predicted rays.

\paragraph{MASt3R~\citep{leroy2025grounding}.}
The official \texttt{MASt3R\_ViTLarge\_BaseDecoder\allowbreak\_512\_catmlpdpt\_metric} (metric-scale) checkpoint at $512$-pixel longest edge with patch size $16$, using DUSt3R-style image normalization (mean and std $0.5$). Pair selection is adaptive on scene size: \texttt{complete} (all pairs) for scenes with $N \le 8$ views, and \texttt{swin-5} (sliding window of $5$) otherwise. The efficiency measurement at $N{=}24$ uses the swin-5 branch ($120$ pairs). The sparse global alignment uses the published defaults: $300$ iterations of coarse alignment at learning rate $0.07$ followed by $300$ iterations of refinement at $0.01$, with per-pixel depth optimization enabled (\texttt{optim\_level=refine+depth}) and matching-confidence threshold $5.0$. Camera intrinsics are shared across views for the single-camera benchmarks (7-Scenes, ScanNet++, nuScenes, DTU) and estimated per-view otherwise (ETH3D).

\paragraph{MASt3R-SfM~\citep{duisterhof2024mast3rsfm}.}
The same MASt3R checkpoint paired with the official training-free retrieval model (\texttt{MASt3R\_ViTLarge\_BaseDecoder\_512\_catmlpdpt\_metric\_retrieval\_trainingfree}). Scene graphs are built via top-$20$ retrieval anchors with top-$10$ retrieved neighbors per anchor (\texttt{retrieval-20-10}), yielding $273$ pairs at $N{=}24$. The sparse global alignment uses the same hyperparameters as MASt3R, including the same per-dataset shared-intrinsics policy.

\FloatBarrier
\section{Analysis of Shared Iterative Refinement}
We examine how the shared model learns over training and how its attention evolves over recurrent iterations.

\subsection{Training Progress}
\label{sec:temporary_training_progress}
\begin{figure}[htbp]
    \centering
    \includegraphics[width=\linewidth]{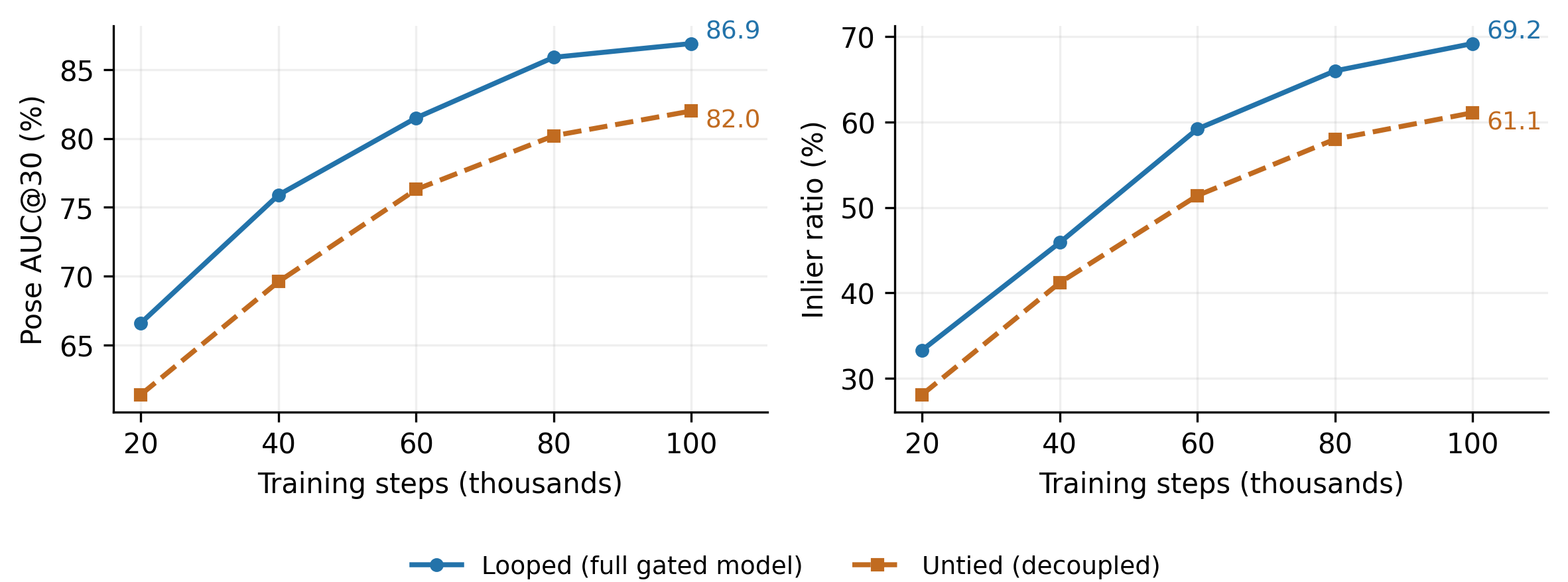}
    \caption{\textbf{Training progress.}
    Pose AUC@$30^\circ$ and inlier ratio at five checkpoints between 20K and 100K training steps. Lines connect the measured checkpoints. The looped model uses the full gated variant in \cref{tab:diag_block}; its advantage over the decoupled baseline persists throughout the reported training interval.}
    \label{fig:temporary_training_progress}
\end{figure}

\cref{fig:temporary_training_progress} compares the looped model with the decoupled baseline at five checkpoints from 20K to 100K training steps. The looped model remains ahead at every reported checkpoint. The AUC@$30^\circ$ gap is 4.9 percentage points at 100K, and the inlier-ratio gap grows from 5.2 points at 20K to 8.1 points at 100K. The advantage therefore persists across the observed training budgets; these checkpoints do not establish the eventual convergence of either model.

\subsection{Emergent Iterative Correspondence Search}
This section extends the correspondence analysis in \cref{sec:analysis} with probing details, an additional visualization, and results for all iterations and individual heads.

\label{sec:emergent_correspondence}

\begin{figure*}[t]
    \centering
    \includegraphics[width=\linewidth]{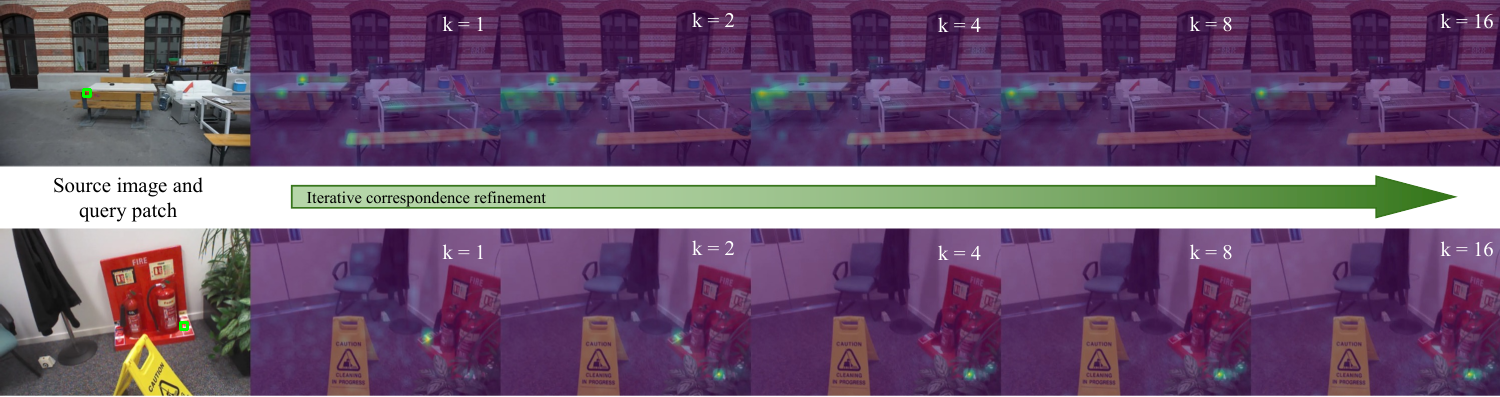}
    \caption{
        \textbf{Emergent iterative correspondence search.}
        For two example query patches (\textcolor{green}{green square}) 
        we visualize the head-averaged global self-attention
        sub-block weights induced by that query at each iteration of
        the loop, overlaid on the corresponding target view.
        Iterations advance left-to-right, the attention starts diffuse
        and progressively concentrates on the corresponding counterpart of
        the queried patch, despite the absence of any explicit feature matching
        supervision during training.
    }
    \label{fig:feature_correspondence}
\end{figure*}

We probe the global attention sub-block of our recurrent layer to investigate how its attention pattern evolves across iterations. For a query patch in view~0, at each iteration $t$ we read the per-head queries and keys $Q^{(t)}_h, K^{(t)}_h$ post $q$ -/ $k$ -LayerNorm and visualize the head-averaged scaled-dot-product attention weights
\[
    \bar a^{(t)} \;=\;
    \tfrac{1}{H} \sum_{h=1}^{H} \operatorname{softmax}\!\left(
        q^{(t)}_h\, K^{(t)\top}_h \big/ \sqrt{d_h}
    \right),
\]
sliced to the patch tokens of each target view and shown as a $H_p \times W_p$ heatmap. 
\cref{fig:feature_correspondence} provides an additional visualization of how attention starts diffuse and progressively concentrates on the corresponding counterpart of the queried patch, correctly resolving symmetries and attending to all geometrically equivalent patches.

\paragraph{Quantitative correspondence protocol.}
We warp query-patch centers using ground-truth depth and relative pose, retaining co-visible queries that pass an occlusion check. For each head and iteration, we restrict the post-softmax global-attention row to the target view's patch tokens and compare its argmax with the ground-truth patch, following~\citet{bratulic2026geometric}. Following the measurement convention of~\citet{stary2025understanding}, we report hit@1 using a Euclidean distance threshold of one patch width. We use the same best-performing head throughout the iteration curve, alongside nearest-neighbor matching on the input DINOv2 features and random selection. The complete iteration sequence and final per-head results are given below.
\begin{table}[htbp]\centering\small
\caption{Best-head correspondence hit@1 (\%), with one-patch tolerance, over all 16 iterations.}
\label{tab:correspondence_full}
\begin{tabular}{@{}lrr@{}}\toprule
Iteration / reference & MegaDepth-1500 & ScanNet-1500 \\ \midrule
Chance & 0.9 & 0.9 \\
DINOv2 nearest neighbor & 49.3 & 40.2 \\
1 & 44.6 & 42.1 \\
2 & 57.8 & 51.3 \\
3 & 65.7 & 58.2 \\
4 & 71.2 & 62.5 \\
5 & 74.5 & 64.8 \\
6 & 76.4 & 66.1 \\
7 & 77.7 & 66.8 \\
8 & 78.4 & 67.2 \\
9 & 78.9 & 67.4 \\
10 & 79.2 & 67.6 \\
11 & 79.4 & 67.7 \\
12 & 79.5 & 67.8 \\
13 & 79.6 & 67.9 \\
14 & 79.7 & 68.1 \\
15 & 79.7 & 68.2 \\
16 & 79.8 & 68.3 \\
\bottomrule\end{tabular}\end{table}

\begin{table}[htbp]\centering\small
\caption{Final-iteration correspondence hit@1 (\%) by attention head. The remaining seven heads are reported as ranges.}
\label{tab:correspondence_heads}
\begin{tabular}{@{}lrrrrrr@{}}\toprule
Dataset & $h_0$ & $h_9$ & $h_4$ & $h_8$ & $h_2$ & Other 7 \\ \midrule
MegaDepth-1500 & 79.8 & 65.3 & 64.2 & 61.1 & 43.9 & 1--11 \\
ScanNet-1500 & 68.3 & 50.8 & 49.1 & 49.0 & 34.5 & 1--9 \\
\bottomrule\end{tabular}\end{table}

\FloatBarrier
\section{Training and Inference Step Budgets}
We first vary the number of steps used in training, then examine how sampling and inference schedules affect the budgets supported by one model. Finally, we study extrapolation beyond the trained range.

\subsection{Fixed-Step Specialists}
\label{sec:additional_ablations}

\paragraph{Fixed-step specialists.} We train and evaluate separate models at fixed $K\in\{8,12,16,24\}$, with no variable-$K$ sampling. All specialists use the same 100K-step ablation protocol and batch size.
Metrics in \cref{tab:fixed_specialists} are averaged equally over DTU, ETH3D, 7-Scenes, ScanNet++, and nuScenes, matching the benchmark set and averaging in \cref{tab:diag_iterative}. Increasing $K$ from 8 to 16 improves every metric. At 24, AUC@$3^\circ$ improves further, while the remaining metrics plateau or slightly worsen. These are separately trained specialists; varying $K$ also changes compute per training example.

\begin{table}[!htbp]\centering\small
\caption{Fixed-step specialist sweep. AUC and IR are percentages ($\uparrow$); both error metrics are lower-is-better.}
\label{tab:fixed_specialists}
\begin{tabular}{@{}lrrrrr@{}}\toprule
$K$ & AUC@$3^\circ$ & AUC@$30^\circ$ & IR@3\% & AbsRel & Rel.~L2 \\ \midrule
8 & 24.3 & 82.8 & 61.7 & 0.060 & 0.052 \\
12 & 30.4 & 85.5 & 67.6 & \underline{0.055} & 0.044 \\
16 & \underline{33.8} & \textbf{86.8} & \textbf{69.6} & \textbf{0.053} & \textbf{0.041} \\
24 & \textbf{35.3} & \underline{86.2} & \underline{69.3} & \textbf{0.053} & \underline{0.042} \\
\bottomrule\end{tabular}\end{table}

The limited gains beyond 16 steps suggest diminishing returns from increasing recurrence alone under this training setup. The best step count can depend on the task and training resources; the results do not establish an architectural ceiling at 16. Jointly increasing training data and compute may allow deeper recurrence to remain useful, but that possibility is not tested by this sweep.

\subsection{Sensitivity to the Training Step Sampler}
\label{sec:sampler_sensitivity}

We vary both the distribution used to sample $K$ (Beta$(2,1)$ or uniform) and its range ($[8,16]$ or $[2,16]$). Table~\ref{tab:sampler_sensitivity} reports the resulting inference-budget trade-off. Uniform sampling on $[2,16]$ gives the strongest results at $K_{\inf}=2$ and $4$, while allocating less training to larger budgets reduces its performance at $12$ and $16$. Beta$(2,1)$ on $[2,16]$ improves AUC@$3^\circ$ at $8$--$16$ relative to our default sampler. Thus, the sampler controls which inference budgets receive more training; the default is not uniformly optimal.

\begin{table}[!htbp]\centering\small
\caption{Sensitivity to the training sampler across inference budgets. Pose AUC and world-space IR at 3\% are percentages (higher is better). $K_{\inf}=2,4$ are outside the training range only for the $[8,16]$ samplers.}
\label{tab:sampler_sensitivity}
\begin{tabular}{@{}lrrrrr@{}}\toprule
Training sampler & $K_{\inf}=2$ & $4$ & $8$ & $12$ & $16$ \\ \midrule
\multicolumn{6}{c}{Pose AUC@$3^\circ$} \\
Beta$(2,1)$, $[8,16]$ & 0.2 & 3.5 & 21.2 & 29.6 & \underline{33.3} \\
Beta$(2,1)$, $[2,16]$ & \underline{1.1} & \underline{9.4} & \textbf{26.3} & \textbf{32.2} & \textbf{34.6} \\
Uniform, $[8,16]$ & 0.3 & 4.4 & \underline{23.9} & \underline{30.9} & 32.4 \\
Uniform, $[2,16]$ & \textbf{3.4} & \textbf{11.9} & 23.0 & 26.5 & 27.1 \\
\midrule
\multicolumn{6}{c}{Pose AUC@$30^\circ$} \\
Beta$(2,1)$, $[8,16]$ & 15.8 & 54.7 & 79.8 & 85.4 & \textbf{86.9} \\
Beta$(2,1)$, $[2,16]$ & \underline{36.5} & \underline{67.6} & \textbf{82.3} & \underline{85.5} & \underline{86.5} \\
Uniform, $[8,16]$ & 18.9 & 60.2 & \underline{82.0} & \textbf{85.6} & \underline{86.5} \\
Uniform, $[2,16]$ & \textbf{52.6} & \textbf{72.1} & 81.3 & 83.3 & 83.8 \\
\midrule
\multicolumn{6}{c}{Inlier ratio at 3\%} \\
Beta$(2,1)$, $[8,16]$ & 1.2 & 24.4 & 58.0 & \textbf{66.7} & \textbf{69.2} \\
Beta$(2,1)$, $[2,16]$ & \underline{11.2} & \underline{42.2} & 60.5 & \underline{66.0} & \underline{67.9} \\
Uniform, $[8,16]$ & 6.7 & 33.0 & \underline{60.6} & 65.8 & \underline{67.9} \\
Uniform, $[2,16]$ & \textbf{26.6} & \textbf{47.9} & \textbf{61.1} & 64.1 & 64.8 \\
\bottomrule\end{tabular}\end{table}

\subsection{Scaling Below $K_{\max}$}
\begin{figure*}[t]
    \centering
    \includegraphics[width=\linewidth]{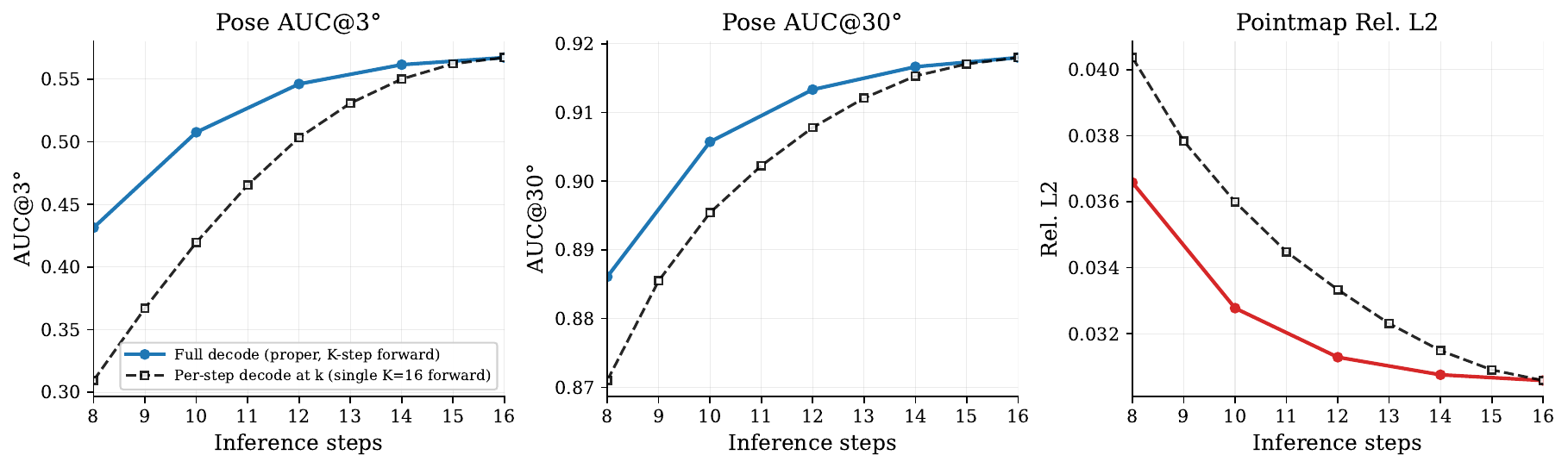}
    \caption{
        \textbf{Decoding intermediate results vs. using lower $K_{\inf}$.}
        We compare using $K_{\inf} < K_{\max}$ steps explicitly to decoding $\mathbf{z}_k$ at intermediate iterations $k$ when $K_{\inf} = K_{\max}$, and show that explicitly conditioning the model on fewer iterations degrades performance less than early-stopping. 
    }
    \label{fig:kinf_intermediate}
\end{figure*}

A sub-$K_{\max}$ compute allows two strategies: a full $K_{\inf}$-step forward
with uniform time interval conditioning calibrated for that budget, or early-stopping a
$K_{\max}$-step rollout by reading $\mathbf{z}_k$ at $k=K_{\inf}$.
\cref{fig:kinf_intermediate} shows the full pass strictly dominates
early-stopping for every $k<K_{\max}$, with the largest gap at the smallest
budget. At $K_{\inf}=8$, Pose~AUC@$3^\circ$ rises from $0.31$ to $0.44$
($\sim\!43\%$ relative improvement).

\subsection{Scaling Beyond $K_{\max}$}
\label{sec:beyond_kmax}

\begin{figure*}[t]
    \centering
    \includegraphics[width=\linewidth]{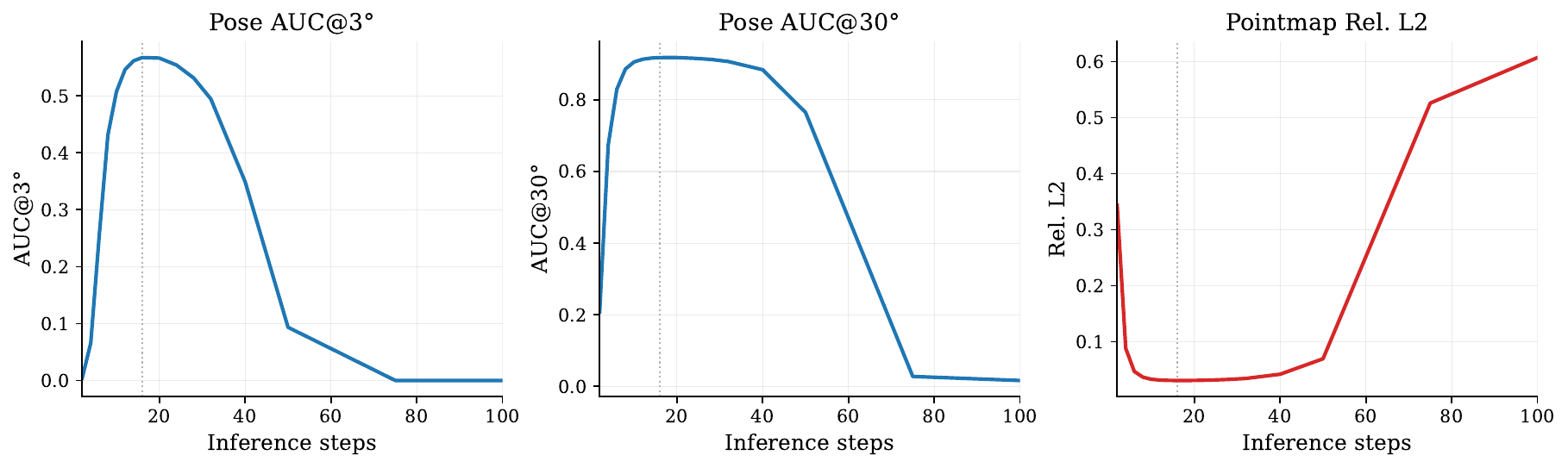}
    \caption{
        \textbf{Step-count extrapolation.} Test-time metrics as a function
        of the inference step count $K_{\inf}$. Performance peaks
        at the maximum trained budget, then degrades when moving far outside the trained range.
    }
    \label{fig:beyond_kmax}
\end{figure*}

DéjàView is trained with an iteration count sampled per-batch as $K \sim \mathrm{Beta}(2,1)$
scaled to $[K_{\min}, K_{\max}]$ with $K_{\max}=16$. We sweep $K_{\inf}$ at test time beyond $K_{\max}$ and observe that Pose~AUC@$3^\circ$ peaks
near the trained budget, then starts to degrade after. Pose~AUC@$30^\circ$ and Pointmap Rel.~L2
remain stable up to approximately $K_{\inf}=30$ but eventually collapse. \cref{fig:beyond_kmax} shows that model iterations cannot be pushed arbitrarily far. 

Our analysis shows that this occurs because some feature channels grow unbounded as we scale beyond $K_{\max}$. The mechanism is already visible inside the trained range (\cref{fig:refinement}b): while cosine similarity to
$\mathbf{z}_{K_{\max}}$ saturates near $1$ and the relative feature update decays, the state norm $\|\mathbf{z}_k\|_2$ keeps growing monotonically through $K_{\max}$ after a short initial contraction. Iterating beyond $K_{\max}$ simply extrapolates this persistent drift: a handful of channels grow without bounds, producing the observed collapse.

\paragraph{Input-injection experiments.}
Following \citet{yang2024looped}, we tested $\mathbf z_{k+1}=f(\mathbf z_k+\mathbf P)$, where $\mathbf P$ is the initial DINOv2 patch embedding, and two variants with a learnable per-channel scale on $\mathbf P$, initialized to zero or one. These variants left within-range performance unchanged and did not prevent degradation beyond $K_{\max}$. Input injection alone did not stabilize our model.

\FloatBarrier
\section{Training Recipe Ablations}
We examine the depth-training stages and the effect of finetuning the image encoder.

\subsection{Two-Stage Depth Training}
\label{sec:suppl_two_stage}

\begin{figure}[t]
    \centering
    \begin{tabular}{@{}c@{\hspace{0.02\linewidth}}c@{\hspace{0.02\linewidth}}c@{}}
        \textbf{Input} & \textbf{Linear Head} & \textbf{Convolutional Head} \\[2pt]
        \includegraphics[width=0.32\linewidth]{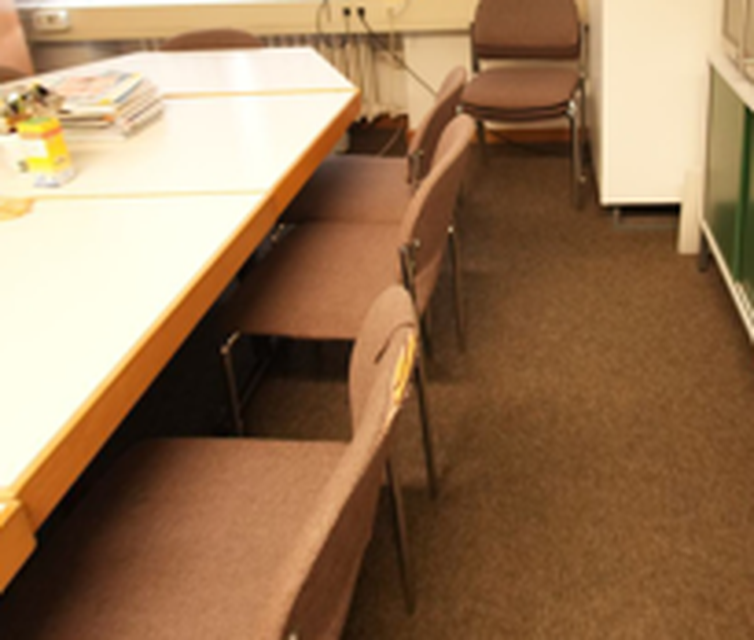} &
        \includegraphics[width=0.32\linewidth]{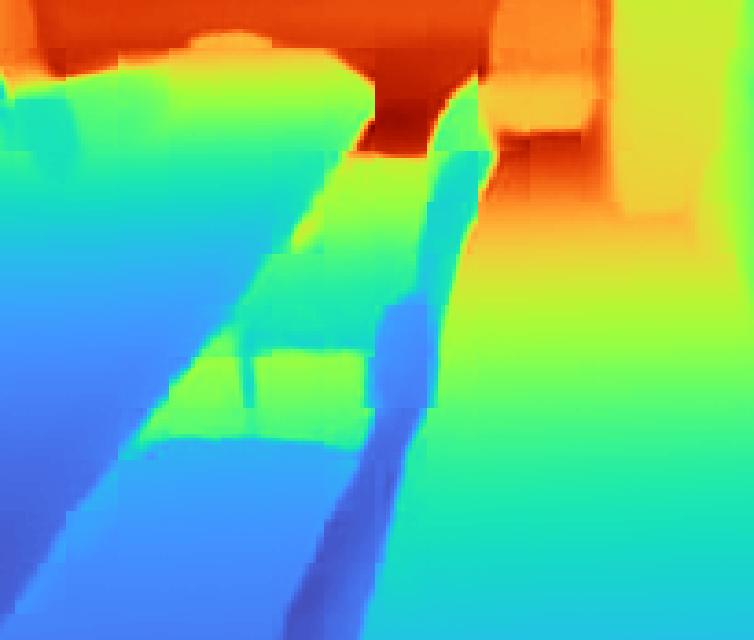} &
        \includegraphics[width=0.32\linewidth]{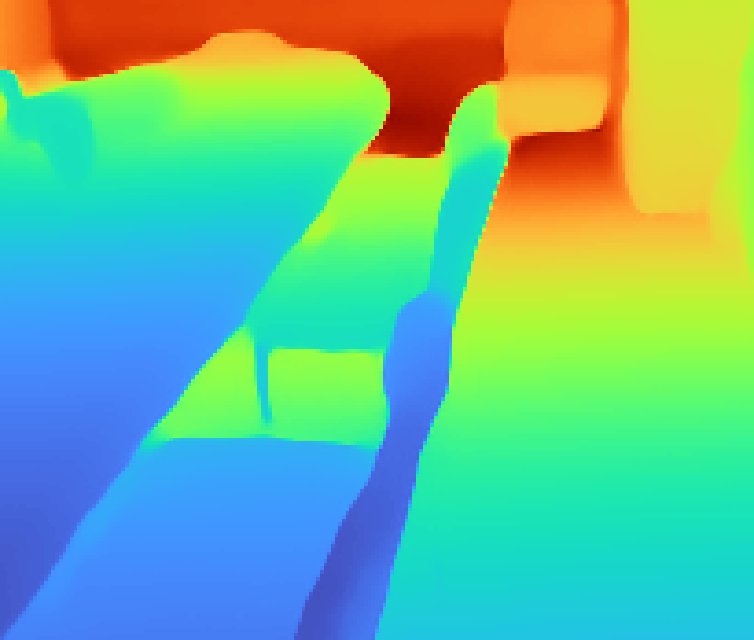}
    \end{tabular}
    \caption{%
        \textbf{Block artifacts.}
        Depth predicted by the first-stage linear head \textbf{(center)} shows visible block artifacts aligned with the DINOv2 patch grid. The second-stage finetune with the convolutional head \textbf{(right)} eliminates them and improves Inlier Ratio.
    }
    \label{fig:suppl_grid_artifacts}
\end{figure}

The linear pixel-shuffle head maps patch tokens to per-pixel depth via pixel shuffling. For ray directions, this is acceptable as patch-border gradients are within $\sim$10\% of intra-patch values. For depth, the same gradients are roughly an order of magnitude larger, producing visible block artifacts at patch boundaries (\cref{fig:suppl_grid_artifacts}).

We address this with two-stage training (\cref{sec:method_training}). The first stage trains the model end-to-end with a linear pixel-shuffle depth head and plain $\ell_2$ depth loss. Training the final pipeline configuration (convolutional head with confidence loss) end-to-end instead yields worse metrics across our benchmarks (\cref{tab:suppl_two_stage_ablation}). The second stage swaps it in, freezes the rest of the network, and finetunes the depth decoder with a confidence-weighted loss. Its convolutions smooth across patch boundaries, eliminating the block pattern (\cref{fig:suppl_grid_artifacts}) and yielding a slight improvement in Inlier Ratio (\cref{tab:suppl_full_recipe_two_stage}). As a result, we also obtain a depth confidence channel that can be used downstream to filter regions with uncertain reconstruction.

\begin{table}[t]
    \caption{%
        \textbf{End-to-end vs.\ two-stage training.}
        Training the final pipeline configuration end-to-end (top, convolutional head with confidence loss) consistently underperforms the first stage of our recipe (bottom, linear head with $\ell_2$ depth loss) on every metric.
    }
    \label{tab:suppl_two_stage_ablation}
    \footnotesize   
    \setlength{\tabcolsep}{6pt}

    \noindent\makebox[\linewidth][c]{%
    \begin{tabular}{l rrrr}
        \toprule
        \textbf{Variant}
            & \textbf{AUC@$3^{\circ}$}~$\uparrow$
            & \textbf{AUC@$30^{\circ}$}~$\uparrow$
            & \textbf{IR}~$\uparrow$
            & \textbf{AbsRel}~$\downarrow$ \\
        \midrule
        Convolutional head + confidence loss (end-to-end) & 23.0 & 77.1 & 56.5 & 0.152 \\
        \midrule
        \textbf{Linear head + $\ell_2$ (our stage 1)}     & \textbf{31.0} & \textbf{80.6} & \textbf{59.2} & \textbf{0.125} \\
        \bottomrule
    \end{tabular}}
\end{table}

\begin{table}[t]
    \caption{%
        \textbf{Stage 2 finetuning.} Finetuning our stage 1 model (linear head + $\ell_2$ depth loss) with the convolutional head and confidence-weighted loss (stage 2) yields a small, consistent gain in Inlier Ratio while leaving the pose metrics unchanged.
    }
    \label{tab:suppl_full_recipe_two_stage}
    \footnotesize
    \setlength{\tabcolsep}{6pt}

    \noindent\makebox[\linewidth][c]{%
    \begin{tabular}{l rrrr}
        \toprule
        \textbf{Variant}
            & \textbf{AUC@$3^{\circ}$}~$\uparrow$
            & \textbf{AUC@$30^{\circ}$}~$\uparrow$
            & \textbf{IR}~$\uparrow$
            & \textbf{AbsRel}~$\downarrow$ \\
        \midrule
        Linear head + $\ell_2$ (stage 1, full recipe) & \textbf{56.8} & \textbf{91.8} & 79.8 & \textbf{0.031} \\
        \midrule
        \textbf{Conv. head + conf. loss (stage 2 finetune)} & \textbf{56.8} & \textbf{91.8} & \textbf{80.3} & \textbf{0.031} \\
        \bottomrule
    \end{tabular}}
\end{table}

\subsection{Encoder Finetuning}
\label{sec:frozen_encoder}
We compare freezing the DINOv2 encoder with finetuning it at $0.1\times$ the base learning rate under the 100K-step ablation protocol. As shown in \cref{tab:frozen_encoder}, freezing reduces pose accuracy and inlier ratio and increases pointmap Rel.~L2 at both inference budgets. This comparison motivates encoder finetuning in our training recipe.
\begin{table}[!htbp]\centering\small
\caption{Encoder finetuning under the 100K-step ablation protocol. The finetuned reference is the variable-$K$ model in \cref{tab:diag_iterative}.}
\label{tab:frozen_encoder}
\begin{tabular}{lrrrrr}\toprule
Encoder & $K_{\inf}$ & AUC@3 $\uparrow$ & AUC@30 $\uparrow$ & IR $\uparrow$ & Rel.~L2 $\downarrow$ \\ \midrule
$0.1\times$ LR & 12 & \underline{29.6} & \underline{85.4} & \underline{66.7} & \underline{0.043} \\
Frozen & 12 & 22.8 & 81.6 & 58.3 & 0.051 \\ \midrule
$0.1\times$ LR & 16 & \textbf{33.3} & \textbf{86.9} & \textbf{69.2} & \textbf{0.040} \\
Frozen & 16 & 24.5 & 82.8 & 60.4 & 0.049 \\
\bottomrule\end{tabular}\end{table}

\FloatBarrier
\section{Additional Reconstruction Evaluation}
These experiments complement the main pointmap and pose benchmarks with surface metrics and evaluation on dynamic sequences.

\subsection{Depth and Surface Reconstruction Metrics}
\label{sec:surface_metrics}

We supplement the pointmap evaluation with depth AbsRel and surface metrics on DTU, ETH3D, and 7-Scenes. Depth AbsRel follows the main evaluation protocol. For surface evaluation, we unproject all evaluated frames of each scene into point clouds and align prediction and ground truth using paired Sim(3) alignment followed by point-to-point ICP refinement, following the protocol in~\cite{wang2025pi}. All methods use the same evaluation pipeline. We report directional nearest-neighbor distances: accuracy measures prediction-to-ground-truth distance, and completeness measures ground-truth-to-prediction distance. We report the mean and median in each direction, without thresholding these distances. The correspondence threshold used during ICP alignment is separate from these metrics. Normal consistency (N.C.) uses the absolute dot product of estimated normals at nearest-neighbor pairs, averaging the two directional means or medians. Depth AbsRel is the mean absolute depth error divided by ground-truth depth over valid pixels.

\begin{table}[!htbp]\centering\footnotesize\setlength{\tabcolsep}{4pt}
\caption{\textbf{DTU depth and surface metrics.} The best, second, and third results use the same colours as \cref{tab:comparison_sota_pmap}. Accuracy and completeness are in mm ($\downarrow$); AbsRel ($\downarrow$); N.C.\ is higher-is-better.}
\label{tab:surface_DTU}
\begin{tabular}{@{}lrrrrrrr@{}}\toprule
\multirow{2}{*}{\textbf{Method}} & \multirow{2}{*}{AbsRel $\downarrow$} & \multicolumn{2}{c}{\textbf{Accuracy} $\downarrow$} & \multicolumn{2}{c}{\textbf{Completeness} $\downarrow$} & \multicolumn{2}{c}{\textbf{N.C.} $\uparrow$} \\
\cmidrule(lr){3-4}\cmidrule(lr){5-6}\cmidrule(lr){7-8}
 & & Mean & Median & Mean & Median & Mean & Median \\ \midrule
DA3-Large & \cellcolor{tabred}0.009 & \cellcolor{tabyellow}0.916 & \cellcolor{tabyellow}0.350 & \cellcolor{tabred}1.140 & \cellcolor{tabyellow}0.414 & \cellcolor{tabyellow}0.584 & \cellcolor{tabyellow}0.631 \\
DA3-Giant & \cellcolor{tabred}0.009 & 1.160 & 0.431 & 1.470 & 0.509 & \cellcolor{taborange}0.591 & \cellcolor{taborange}0.641 \\
VGGT & \cellcolor{tabred}0.009 & 0.935 & 0.477 & 1.810 & 1.030 & \cellcolor{tabred}0.615 & \cellcolor{tabred}0.677 \\
Pi3 & \cellcolor{tabyellow}0.012 & \cellcolor{taborange}0.800 & \cellcolor{taborange}0.346 & \cellcolor{taborange}1.170 & \cellcolor{tabred}0.313 & 0.578 & 0.622 \\
\midrule
\textbf{Ours} & \cellcolor{taborange}0.010 & \cellcolor{tabred}0.796 & \cellcolor{tabred}0.342 & \cellcolor{tabyellow}1.340 & \cellcolor{taborange}0.407 & 0.580 & 0.623 \\
\bottomrule\end{tabular}\end{table}

\begin{table}[!htbp]\centering\footnotesize\setlength{\tabcolsep}{4pt}
\caption{ETH3D depth and surface metrics. Accuracy and completeness are in m ($\downarrow$); AbsRel  ($\downarrow$); N.C.\ is higher-is-better.}
\label{tab:surface_ETH3D}
\begin{tabular}{@{}lrrrrrrr@{}}\toprule
\multirow{2}{*}{\textbf{Method}} & \multirow{2}{*}{AbsRel $\downarrow$} & \multicolumn{2}{c}{\textbf{Accuracy} $\downarrow$} & \multicolumn{2}{c}{\textbf{Completeness} $\downarrow$} & \multicolumn{2}{c}{\textbf{N.C.} $\uparrow$} \\
\cmidrule(lr){3-4}\cmidrule(lr){5-6}\cmidrule(lr){7-8}
 & & Mean & Median & Mean & Median & Mean & Median \\ \midrule
DA3-Large & 0.035 & 0.226 & 0.146 & 0.208 & 0.102 & 0.783 & 0.888 \\
DA3-Giant & \cellcolor{tabyellow}0.023 & \cellcolor{tabyellow}0.165 & 0.094 & 0.112 & \cellcolor{tabyellow}0.054 & 0.831 & 0.922 \\
VGGT & 0.037 & \cellcolor{taborange}0.148 & \cellcolor{tabyellow}0.085 & \cellcolor{tabyellow}0.105 & \cellcolor{taborange}0.051 & \cellcolor{tabyellow}0.833 & \cellcolor{tabyellow}0.951 \\
Pi3 & \cellcolor{taborange}0.021 & \cellcolor{tabred}0.081 & \cellcolor{taborange}0.044 & \cellcolor{tabred}0.047 & \cellcolor{tabred}0.023 & \cellcolor{taborange}0.852 & \cellcolor{tabred}0.966 \\
\midrule
\textbf{Ours} & \cellcolor{tabred}0.018 & \cellcolor{tabred}0.081 & \cellcolor{tabred}0.041 & \cellcolor{taborange}0.048 & \cellcolor{tabred}0.023 & \cellcolor{tabred}0.854 & \cellcolor{taborange}0.953 \\
\bottomrule\end{tabular}\end{table}

\begin{table}[!htbp]\centering\footnotesize\setlength{\tabcolsep}{4pt}
\caption{7-Scenes depth and surface metrics. Accuracy and completeness are in m ($\downarrow$); AbsRel ($\downarrow$); N.C.\ is higher-is-better.}
\label{tab:surface_7-Scenes}
\begin{tabular}{@{}lrrrrrrr@{}}\toprule
\multirow{2}{*}{\textbf{Method}} & \multirow{2}{*}{AbsRel $\downarrow$} & \multicolumn{2}{c}{\textbf{Accuracy} $\downarrow$} & \multicolumn{2}{c}{\textbf{Completeness} $\downarrow$} & \multicolumn{2}{c}{\textbf{N.C.} $\uparrow$} \\
\cmidrule(lr){3-4}\cmidrule(lr){5-6}\cmidrule(lr){7-8}
 & & Mean & Median & Mean & Median & Mean & Median \\ \midrule
DA3-Large & \cellcolor{tabyellow}0.050 & \cellcolor{tabyellow}0.016 & \cellcolor{tabyellow}0.006 & \cellcolor{tabyellow}0.020 & \cellcolor{tabred}0.006 & 0.606 & 0.666 \\
DA3-Giant & \cellcolor{taborange}0.049 & \cellcolor{taborange}0.014 & \cellcolor{taborange}0.005 & \cellcolor{taborange}0.019 & \cellcolor{taborange}0.007 & \cellcolor{tabred}0.614 & \cellcolor{tabred}0.679 \\
VGGT & 0.058 & 0.017 & \cellcolor{taborange}0.005 & 0.024 & \cellcolor{tabyellow}0.010 & \cellcolor{taborange}0.612 & \cellcolor{taborange}0.675 \\
Pi3 & \cellcolor{taborange}0.049 & \cellcolor{tabred}0.011 & \cellcolor{tabred}0.004 & \cellcolor{tabred}0.015 & \cellcolor{taborange}0.007 & \cellcolor{tabyellow}0.611 & \cellcolor{tabyellow}0.674 \\
\midrule
\textbf{Ours} & \cellcolor{tabred}0.047 & \cellcolor{tabred}0.011 & \cellcolor{tabred}0.004 & \cellcolor{taborange}0.019 & \cellcolor{taborange}0.007 & 0.602 & 0.659 \\
\bottomrule\end{tabular}\end{table}

Our model achieves the lowest or tied-lowest Chamfer accuracy in these tables and the lowest depth AbsRel on ETH3D and 7-Scenes. Completeness and normal consistency show a mixed ranking: strong pointwise accuracy does not imply the best result on every surface metric.

\subsection{Evaluation on Dynamic Scenes}
\label{sec:dynamic_results}

We evaluate video depth on Sintel, Bonn, and KITTI and camera trajectories on Sintel and TUM-dynamics, following the evaluation protocol of \cite{wang2025pi}. Video depth uses sequence-level scale-and-shift alignment with a least-absolute-deviation objective; dataset scores are weighted by the number of valid pixels. Tables~\ref{tab:dynamic_depth} and~\ref{tab:dynamic_pose} report the results.

\begin{table}[!htbp]\centering\footnotesize\setlength{\tabcolsep}{4pt}
\caption{Video depth estimation. Each dataset reports AbsRel ($\downarrow$) and the fraction of predictions with depth ratio $\delta<1.25$ ($\uparrow$).}
\label{tab:dynamic_depth}
\begin{tabular}{@{}lrrrrrr@{}}\toprule
\multirow{2}{*}{\textbf{Method}} & \multicolumn{2}{c}{\textbf{Sintel}} & \multicolumn{2}{c}{\textbf{Bonn}} & \multicolumn{2}{c}{\textbf{KITTI}} \\
\cmidrule(lr){2-3}\cmidrule(lr){4-5}\cmidrule(lr){6-7}
 & AbsRel $\downarrow$ & $\delta<1.25\uparrow$ & AbsRel $\downarrow$ & $\delta<1.25\uparrow$ & AbsRel $\downarrow$ & $\delta<1.25\uparrow$ \\ \midrule
DA3-Large & 0.2523 & 0.6697 & 0.0471 & 0.9729 & 0.0537 & 0.9728 \\
DA3-Giant & \cellcolor{tabyellow}0.2122 & \cellcolor{taborange}0.6936 & \cellcolor{tabyellow}0.0466 & 0.9727 & \cellcolor{tabyellow}0.0529 & \cellcolor{taborange}0.9775 \\
VGGT & 0.2258 & 0.6854 & 0.0495 & \cellcolor{tabyellow}0.9732 & 0.0589 & 0.9621 \\
Pi3 & \cellcolor{taborange}0.2044 & \cellcolor{tabred}0.7362 & \cellcolor{tabred}0.0454 & \cellcolor{tabred}0.9745 & \cellcolor{tabred}0.0353 & \cellcolor{tabred}0.9871 \\
\midrule
\textbf{Ours} & \cellcolor{tabred}0.1983 & \cellcolor{tabyellow}0.6922 & \cellcolor{taborange}0.0455 & \cellcolor{taborange}0.9743 & \cellcolor{taborange}0.0415 & \cellcolor{tabyellow}0.9772 \\
\bottomrule\end{tabular}\end{table}

\begin{table}[!htbp]\centering\footnotesize\setlength{\tabcolsep}{4pt}
\caption{Camera trajectory estimation. ATE and translational/rotational RPE are lower-is-better. Trajectories are aligned by Sim(3); ATE is the root-mean-square camera-center error. RPE is the root-mean-square relative-pose error between consecutive evaluated frames, with translation in meters and rotation in degrees.}
\label{tab:dynamic_pose}
\begin{tabular}{@{}lrrrrrr@{}}\toprule
\multirow{2}{*}{\textbf{Method}} & \multicolumn{3}{c}{\textbf{Sintel}} & \multicolumn{3}{c}{\textbf{TUM-dynamics}} \\
\cmidrule(lr){2-4}\cmidrule(lr){5-7}
 & ATE (m) & RPE tr. (m) & RPE rot. ($^\circ$) & ATE (m) & RPE tr. (m) & RPE rot. ($^\circ$) \\ \midrule
DA3-Large & 0.1307 & 0.0628 & 0.8016 & 0.0120 & 0.0104 & 0.6722 \\
DA3-Giant & \cellcolor{taborange}0.0961 & \cellcolor{tabyellow}0.0601 & 0.8211 & \cellcolor{taborange}0.0111 & \cellcolor{tabyellow}0.0089 & 0.6923 \\
VGGT & 0.1715 & 0.0615 & \cellcolor{tabyellow}0.4704 & \cellcolor{tabyellow}0.0118 & 0.0100 & \cellcolor{taborange}0.3094 \\
Pi3 & \cellcolor{tabred}0.0729 & \cellcolor{tabred}0.0373 & \cellcolor{tabred}0.2875 & 0.0140 & \cellcolor{taborange}0.0088 & \cellcolor{tabred}0.3059 \\
\midrule
\textbf{Ours} & \cellcolor{tabyellow}0.1191 & \cellcolor{taborange}0.0429 & \cellcolor{taborange}0.3542 & \cellcolor{tabred}0.0090 & \cellcolor{tabred}0.0082 & \cellcolor{tabyellow}0.3185 \\
\bottomrule\end{tabular}\end{table}

Our model obtains the lowest Sintel depth AbsRel and the lowest TUM ATE and translational RPE among these methods. Pi3 performs better on several other metrics, including both depth metrics on KITTI and all three trajectory metrics on Sintel. Our large training corpus is predominantly static (Section~\ref{sec:training_datasets}), whereas Pi3 also uses an internal dynamic-scene dataset~\citep{wang2025pi}; this difference may contribute to its advantage on some dynamic-scene metrics.

\FloatBarrier
\section{Societal Impact}
\label{sec:suppl_societal_impact}
\methodname{} reconstructs 3D geometry and camera poses from images. While this capability is not new, our work brings strong reconstruction quality within reach at a smaller scale than recent feed-forward baselines, lowering the overall cost of deployment. Because the model outputs geometry rather than photorealistic imagery, the direct risk of deceptive media generation is lower than for image or video synthesis models, though deployments that combine reconstruction with generative rendering should still consider provenance signals such as watermarking~\citep{wen2023treering}. From an environmental perspective, our ViT-B model is trained on 128 H100 GPUs, comparable to other recent feed-forward reconstruction methods. At inference, however, it operates with roughly an order of magnitude fewer parameters and a memory footprint of under 5 GiB at $24$ input views, reducing the per-query resource cost of deployment relative to the larger baselines we evaluate.

\clearpage
\section{Additional Qualitative Results}
\label{sec:additional_qualitative}
\begin{figure}[!ht]
    \centering
    \includegraphics[width=\linewidth]{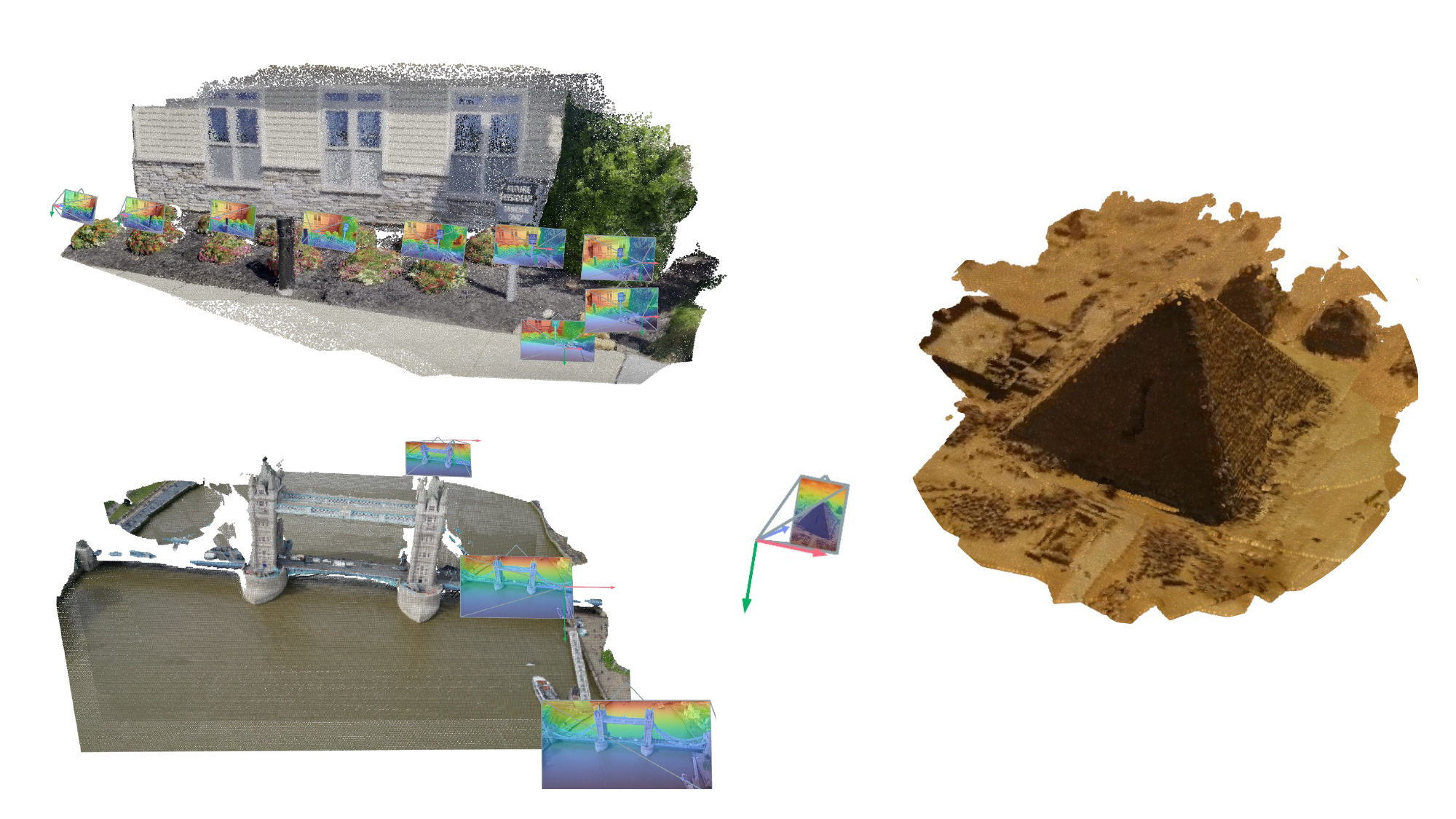}
    \caption{
        \textbf{Qualitative results.}
        Predicted point clouds and cameras for in-the-wild captures.
    }
    \label{fig:qualitative2}
    \vspace{-4mm}
\end{figure}

\clearpage
\section*{NeurIPS Paper Checklist}
\begin{enumerate}
\item \textbf{Claims}
\begin{itemize}
\item[] \textbf{Question:} Do the main claims made in the abstract and introduction accurately reflect the paper's contributions and scope?
\item[] \textbf{Answer:} [Yes]
\item[] \textbf{Justification:} The abstract and Section 1 state our two contributions (a weight-tied recurrent block trained with variable step count K, and the directional-refinement analysis) and the scope of the empirical claim (matching or outperforming larger feed-forward baselines on five reconstruction benchmarks at a fraction of the parameters). The empirical claims are supported by Section 4.1 and the diagnostics in Section 4.2.
\end{itemize}

\item \textbf{Limitations}
\begin{itemize}
\item[] \textbf{Question:} Does the paper discuss the limitations of the work performed by the authors?
\item[] \textbf{Answer:} [Yes]
\item[] \textbf{Justification:} Section 5 discusses the limited extrapolation beyond the trained step range, the trade-off between variable-K flexibility and fixed-K specialist quality, and the lack of an explicit scene-dynamics model.
\end{itemize}

\item \textbf{Theory assumptions and proofs}
\begin{itemize}
\item[] \textbf{Question:} For each theoretical result, does the paper provide the full set of assumptions and a complete (and correct) proof?
\item[] \textbf{Answer:} [N/A]
\item[] \textbf{Justification:} The paper presents an empirical contribution and does not include formal theorems or proofs.
\end{itemize}

\item \textbf{Experimental result reproducibility}
\begin{itemize}
\item[] \textbf{Question:} Does the paper fully disclose all the information needed to reproduce the main experimental results of the paper to the extent that it affects the main claims and/or conclusions of the paper (regardless of whether the code and data are provided or not)?
\item[] \textbf{Answer:} [Yes]
\item[] \textbf{Justification:} Section 3 fully specifies the architecture (recurrent block, decoder heads, time conditioning, scale vectors), Section 4 lists training datasets, optimizer, learning rate schedule, batch composition, and step counts, and Section~\ref{sec:suppl_baselines} details every baseline configuration used in the comparison. Section~\ref{sec:suppl_two_stage} additionally documents the two-stage training rationale.
\end{itemize}

\item \textbf{Open access to data and code}
\begin{itemize}
\item[] \textbf{Question:} Does the paper provide open access to the data and code, with sufficient instructions to faithfully reproduce the main experimental results, as described in supplemental material?
\item[] \textbf{Answer:} [Yes]
\item[] \textbf{Justification:} Code, evaluation instructions, and training configurations are publicly available at \url{https://github.com/nv-tlabs/dvlt}, and trained weights are available at \url{https://huggingface.co/nvidia/dvlt}. The repository documents installation, dataset preparation, and evaluation on the five main benchmarks. Training datasets are available from their original providers and are listed in Section~\ref{sec:training_datasets}.
\end{itemize}

\item \textbf{Experimental setting/details}
\begin{itemize}
\item[] \textbf{Question:} Does the paper specify all the training and test details (e.g., data splits, hyperparameters, how they were chosen, type of optimizer) necessary to understand the results?
\item[] \textbf{Answer:} [Yes]
\item[] \textbf{Justification:} Section 4 specifies the optimizer (AdamW), learning rates and schedules for both training stages, weight decay, loss weights, the variable-K sampling distribution, training step counts, image resolution, and view-count range. Evaluation datasets, metrics, and the Sim(3) alignment protocol are specified in Section 4.1, and per-baseline checkpoints and configurations are in Section~\ref{sec:suppl_baselines}.
\end{itemize}

\item \textbf{Experiment statistical significance}
\begin{itemize}
\item[] \textbf{Question:} Does the paper report error bars suitably and correctly defined or other appropriate information about the statistical significance of the experiments?
\item[] \textbf{Answer:} [No]
\item[] \textbf{Justification:} We report results from single training runs and do not include error bars over training seeds. Multi-seed runs are computationally impractical for the main comparisons. The diagnostics in Section 4.2 use a shared training protocol so that variants are directly comparable.
\end{itemize}

\item \textbf{Experiments compute resources}
\begin{itemize}
\item[] \textbf{Question:} For each experiment, does the paper provide sufficient information on the computer resources (type of compute workers, memory, time of execution) needed to reproduce the experiments?
\item[] \textbf{Answer:} [Yes]
\item[] \textbf{Justification:} Section 4 states that training uses 128 H100 GPUs with the specified per-step token and image budgets. Table 3 reports inference compute (TFLOPs), per-image compute, and peak inference memory measured on a single A100 with 24 input views. Section~\ref{sec:suppl_societal_impact} also discusses training compute.
\end{itemize}

\item \textbf{Code of ethics}
\begin{itemize}
\item[] \textbf{Question:} Does the research conducted in the paper conform, in every respect, with the NeurIPS Code of Ethics \url{https://neurips.cc/public/EthicsGuidelines}?
\item[] \textbf{Answer:} [Yes]
\item[] \textbf{Justification:} The work uses publicly available academic datasets and develops a 3D reconstruction model. We have reviewed the NeurIPS Code of Ethics and the research conforms with it.
\end{itemize}

\item \textbf{Broader impacts}
\begin{itemize}
\item[] \textbf{Question:} Does the paper discuss both potential positive societal impacts and negative societal impacts of the work performed?
\item[] \textbf{Answer:} [Yes]
\item[] \textbf{Justification:} Section~\ref{sec:suppl_societal_impact} discusses the positive impact of lowering the resource cost of 3D reconstruction at inference, the relatively limited deceptive-media risk because the model outputs geometry rather than photorealistic imagery, and the implications for downstream pipelines that combine reconstruction with generative rendering.
\end{itemize}

\item \textbf{Safeguards}
\begin{itemize}
\item[] \textbf{Question:} Does the paper describe safeguards that have been put in place for responsible release of data or models that have a high risk for misuse (e.g., pre-trained language models, image generators, or scraped datasets)?
\item[] \textbf{Answer:} [N/A]
\item[] \textbf{Justification:} The model outputs 3D geometry and camera poses rather than photorealistic imagery, so does not present the misuse profile of generative image or video models. The training data consists of publicly released academic 3D reconstruction datasets.
\end{itemize}

\item \textbf{Licenses for existing assets}
\begin{itemize}
\item[] \textbf{Question:} Are the creators or original owners of assets (e.g., code, data, models), used in the paper, properly credited and are the license and terms of use explicitly mentioned and properly respected?
\item[] \textbf{Answer:} [Yes]
\item[] \textbf{Justification:} All baselines, training datasets, evaluation datasets, and the DINOv2 encoder are cited in the corresponding sections of the paper (Sections 3, 4, 4.1 and~\ref{sec:suppl_baselines}). All are publicly released assets used under their published terms of use.
\end{itemize}

\item \textbf{New assets}
\begin{itemize}
\item[] \textbf{Question:} Are new assets introduced in the paper well documented and is the documentation provided alongside the assets?
\item[] \textbf{Answer:} [Yes]
\item[] \textbf{Justification:} The released code is documented in the repository README and accompanying installation, data, and evaluation documentation (\url{https://github.com/nv-tlabs/dvlt}). The released model includes a configuration, model card describing its architecture, training data, intended use and limitations, and license (\url{https://huggingface.co/nvidia/dvlt}). Code is primarily licensed under Apache 2.0, with third-party licenses documented separately; model weights are released under the NVIDIA non-commercial license.
\end{itemize}

\item \textbf{Crowdsourcing and research with human subjects}
\begin{itemize}
\item[] \textbf{Question:} For crowdsourcing experiments and research with human subjects, does the paper include the full text of instructions given to participants and screenshots, if applicable, as well as details about compensation (if any)?
\item[] \textbf{Answer:} [N/A]
\item[] \textbf{Justification:} The paper does not involve crowdsourcing or research with human subjects.
\end{itemize}

\item \textbf{Institutional review board (IRB) approvals or equivalent for research with human subjects}
\begin{itemize}
\item[] \textbf{Question:} Does the paper describe potential risks incurred by study participants, whether such risks were disclosed to the subjects, and whether Institutional Review Board (IRB) approvals (or an equivalent approval/review based on the requirements of your country or institution) were obtained?
\item[] \textbf{Answer:} [N/A]
\item[] \textbf{Justification:} The paper does not involve human subjects.
\end{itemize}

\item \textbf{Declaration of LLM usage}
\begin{itemize}
\item[] \textbf{Question:} Does the paper describe the usage of LLMs if it is an important, original, or non-standard component of the core methods in this research? Note that if the LLM is used only for writing, editing, or formatting purposes and does not impact the core methodology, scientific rigor, or originality of the research, declaration is not required.
\item[] \textbf{Answer:} [N/A]
\item[] \textbf{Justification:} LLMs are not a component of our core methodology.
\end{itemize}

\end{enumerate}

\end{document}